\documentclass[journal]{IEEEtran}

\usepackage{amsmath,amsfonts}
\usepackage{algorithmic}
\usepackage{algorithm}
\usepackage{array}
\usepackage[caption=false,font=normalsize,labelfont=sf,textfont=sf]{subfig}
\usepackage{textcomp}
\usepackage{stfloats}
\usepackage{url}
\usepackage{verbatim}
\usepackage{graphicx}
\usepackage{cite}
\usepackage{booktabs}
\usepackage{multirow}
\usepackage{xcolor}
\usepackage[table]{xcolor}
\newcommand{\zyw}[1]{\textcolor{black}{#1}}
\newcommand{\hb}[1]{\textcolor{black}{#1}}

\begin{document}

\title{Multi-modal Generative AI:\\Multi-modal LLMs, Diffusions, and the Unification}

\author{
Xin~Wang,~\IEEEmembership{Member,~IEEE},~Yuwei~Zhou,~Bin~Huang,~Hong~Chen,~and~Wenwu~Zhu,~\IEEEmembership{Fellow,~IEEE}

\thanks{Xin Wang, Yuwei Zhou, Bin Huang, Hong Chen, and Wenwu Zhu are with the Department of Computer Science, Beijing Information Science and Technology National Research Center, Tsinghua University, Beijing 100084, China. (E-mail: \{xin\_wang, wwzhu\}@tsinghua.edu.cn), \{zhou-yw21, huangb23, h-chen20\}@mails.tsinghua.edu.cn.}
\thanks{Corresponding author: Wenwu Zhu}
\thanks{This work was supported by the National Natural Science Foundation of China No. 62222209, Beijing National Research Center for Information Science and Technology under Grant No. BNR2023TD03006, and Beijing Key Lab of Networked Multimedia.}
}

\markboth{IEEE Transactions on Circuits and Systems for Video Technology}%
{Shell \MakeLowercase{\textit{Xin Wang et al.}}: Multi-modal Generative AI: Multi-modal LLMs, Diffusions and the Unification}


\maketitle

\begin{abstract}
Multi-modal generative AI (Artificial Intelligence) has attracted increasing attention from both academia and industry. Particularly, two dominant families of techniques have emerged: i) Multi-modal large language models (LLMs) demonstrate impressive ability for \textit{multi-modal understanding}; and ii) Diffusion models exhibit remarkable multi-modal powers in terms of \textit{multi-modal generation}. Therefore, this paper provides a comprehensive overview of multi-modal generative AI, including multi-modal LLMs, diffusions, and the unification for understanding and generation. To lay a solid foundation for unified models, we first provide a detailed review of both multi-modal LLMs and diffusion models, respectively, including their probabilistic modeling procedure, multi-modal architecture design, and advanced applications to image/video LLMs as well as text-to-image/video generation. Furthermore, we explore the emerging efforts toward unified models for understanding and generation. To achieve the unification of understanding and generation, we investigate key designs including autoregressive-based and diffusion-based modeling, as well as dense and Mixture-of-Experts (MoE) architectures. We then introduce several strategies for unified models, analyzing their potential advantages and disadvantages. In addition, we summarize the common datasets widely used for multi-modal generative AI pretraining. Last but not least, we present several challenging future research directions that may contribute to the ongoing advancement of multi-modal generative AI.
\end{abstract}


\section{Introduction}
\label{sec:intro}
Multi-modal generative AI (Artificial Intelligence) has received increasing attention recently with the advent of (multi-modal) large language models (LLMs) and diffusion models. Two typical models of multi-modal generative AI are GPT-4V~\cite{GPT4V} and Sora~\cite{sora} from OpenAI, which have produced great impacts on both academia and industry. To compare GPT-4V and Sora in terms of functionality, GPT-4V targets multi-modal understanding, and Sora aims at visual generation --- GPT-4V enables the LLM to understand visual input via generating relevant texts, while Sora serves as a text-to-video generation model which outputs visual signals given textual input. To make comparisons in terms of probabilistic modeling, GPT-4V is a multi-modal LLM with autoregressive probabilistic modeling, while Sora is a multi-modal video generation model with diffusion denoising modeling. 

As such, there naturally arises a question: ``Is it possible to establish a unified multi-modal generative model for simultaneous understanding and generation?'' And if the answer is \textit{yes}, what would such a model be, either similar to multi-modal LLM or diffusion, or in a new form? To capture the relations among different modalities, is it a good idea to adopt an early-fusion strategy (such as Chameleon~\cite{chameleon}), or just straightforwardly align a pretrained visual model with a language model (such as LLAVA~\cite{llava})? To further unify understanding and generation, is it sufficient to employ Mixture of Experts (MoE) strategies or only use a dense model?

\begin{figure*}
    \centering
    \includegraphics[width=\linewidth]{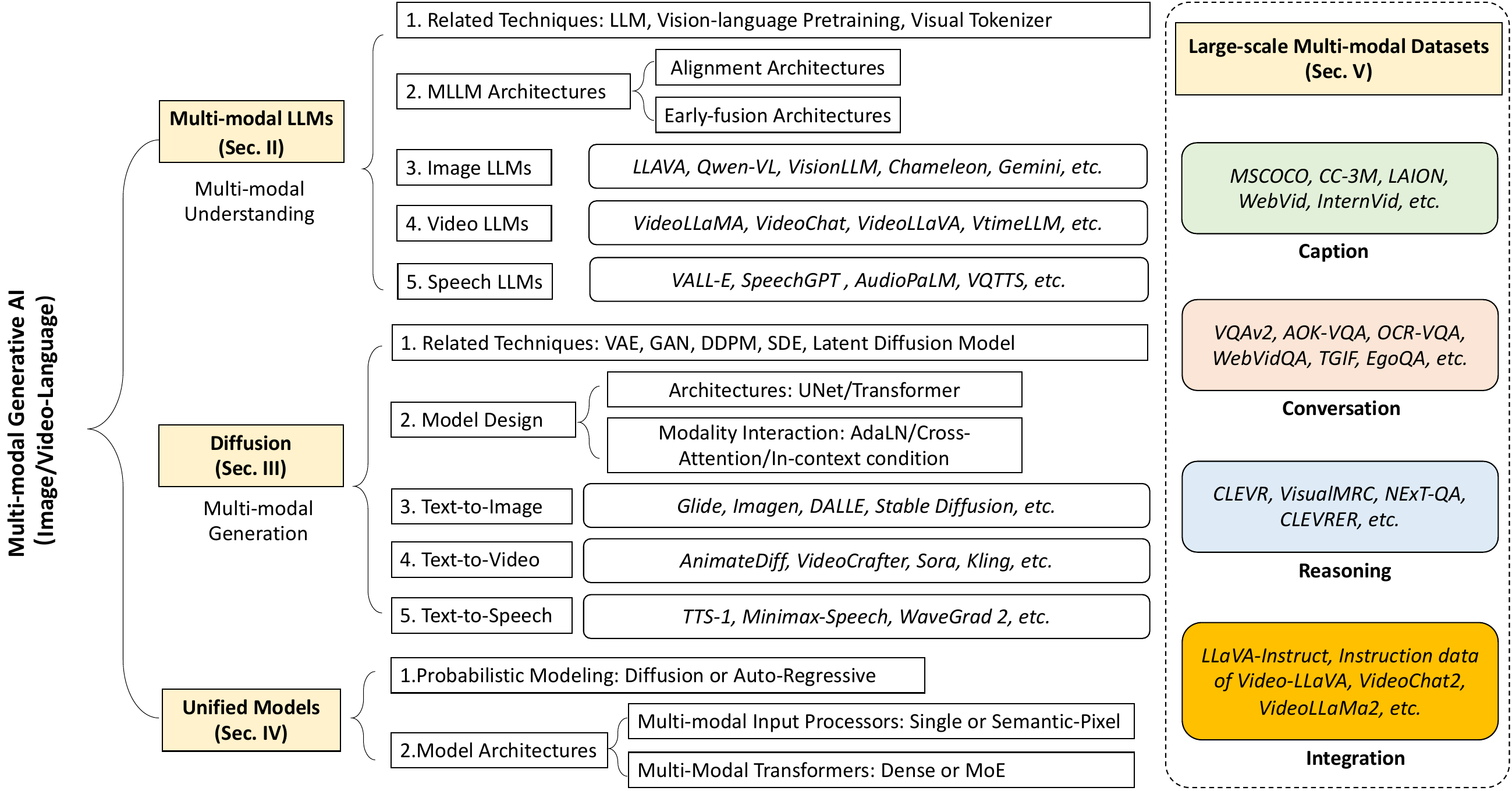}
    \caption{\zyw{The overall organization of this paper.}}
    \label{fig:overall}
\end{figure*}

To answer these questions, we conduct deep and comprehensive discussions of multi-modal generative AI in this paper, whose overall organization is illustrated in Fig.~\ref{fig:overall}. Specifically, we first present a systematic review of existing works on multi-modal LLM (Sec.~\ref{sec:MLLM}) and multi-modal diffusion (Sec.~\ref{sec:diffusion}), covering mathematical preliminaries, model architectures, fusion strategies, recent advances, and applications. Then we present our insights on unified models for simultaneous understanding and generation in Sec.~\ref{sec:unified framework}. Besides, we further summarize video/visual-language datasets for multi-modal generative AI pretraining in Sec.~\ref{sec:data}. Last, we provide future directions that deserve further investigation for multi-modal generative AI.

\zyw{In this paper, our scope primarily lies in multi-modal understanding, generation, and their unification. Some concepts widely studied in the field of LLMs, such as in-context learning, post-training techniques (e.g., supervised fine-tuning and reinforcement learning), sparse attention, and positional embeddings, are important but not the main focus of this survey. Readers interested in these topics are referred to related surveys such as~\cite{minaee2024large,zhao2023survey}. Instead, we focus on recent high-quality works adapted to the multi-modal generative setting, providing a comprehensive overview of the mechanisms that enable multi-modal understanding and generation.}

We would like to point out that although several insightful surveys have been conducted on multi-modal understanding~\cite{liang2024survey,wu2023multimodal,caffagni2024revolution}, visual generation~\cite{croitoru2023diffusion,yang2023diffusion,cao2024survey,nazarieh2024survey,li2024introduction}, and both~\cite{zhang2025unified,xie2024towards}, this work differs from them in comprehensive discussions on models for the unification of understanding and generation in addition to reviewing them separately, thus contributing to the ongoing advancement of multi-modal generative AI. We highlight recent advances, categorize existing approaches, introduce related datasets, and share insights for future directions. In summary, we make the following contributions.

\begin{itemize}
    \item We comprehensively overview multi-modal generative AI, covering multi-modal LLMs for multi-modal understanding and diffusion models for visual generation.
    \item We propose a structured taxonomy of unified models for multi-modal understanding and generation, and provide thorough discussions on them.
    \item We share our insights on promising future directions to highlight the trending research for advances in multi-modal generative AI.
\end{itemize}

\section{Multi-modal LLM for Understanding}
\label{sec:MLLM}
Multi-modal LLMs have recently become dominant in the field of understanding. In this section, we will review the literature on the multi-modal LLMs.

\subsection{Preliminaries}
We first introduce some preliminaries involving the LLM, vision-language pretraining, and visual tokenizers.
\subsubsection{LLM Autoregressive Probabilistic Modeling}
The core component of multi-modal LLMs is the LLM, which receives the multi-modal input, including the user's instructions, questions, and visual information, and then outputs the answers to the user in a text-generation form. The LLM is basically an autoregressive model that tries to predict the next word based on all the previous words, as shown in Eq.~\eqref{eq:ar}.

\begin{equation}
\label{eq:ar}
    p(w) = \prod_{i=1}^n p_{\theta_L}(w_i|w_{<i}),
\end{equation}
where $\theta_L$ denotes the parameters of the LLM, which is generally composed of several layers of transformers~\cite{transformer}. Note that LLM can only receive the text tokens as its input. The next important problem for multi-modal LLM is how to enable LLM to understand the visual information. To tackle the problem, most existing works~\cite{llava, huang2024vtimellm, LLaVA-OneVision} try to align the LLM with the visual encoders from vision-language pretraining tasks, such as CLIP~\cite{CLIP}. More recently, there have been some attempts~\cite{chameleon} to directly transform the images into discrete visual tokens so that the text and visual tokens can be tackled by the autoregressive LLM together. Next, we will introduce preliminaries about vision-language pretraining and visual tokenizers.

\subsubsection{Vision-Language Pretraining}

Vision-language pretraining (VLP) aims to learn aligned representations of images and texts by leveraging large-scale image-text pairs. One of the most influential VLP models is CLIP~\cite{CLIP}, which learns a joint embedding space where semantically related images and texts are mapped close to each other. 

CLIP consists of two separate encoders: a visual encoder (typically a Vision Transformer~\cite{vit} or ResNet~\cite{resnet}) and a text encoder (usually a Transformer). Given a batch of image-text pairs, CLIP is trained with a contrastive loss that encourages the embeddings of matched image-text pairs to be close while pushing apart the embeddings of mismatched pairs.

The pretrained CLIP model has been widely used in multi-modal LLMs to inject visual understanding into LLMs. Typically, visual features extracted by the CLIP image encoder are projected into the input space of LLM through a learned adapter or alignment module~\cite{llava}. This allows LLMs to reason over both linguistic and visual information in a unified manner.

\subsubsection{Visual Tokenizer}\label{sec:visual_tokenizer}
\begin{figure}
    \centering
    \includegraphics[width=\linewidth]{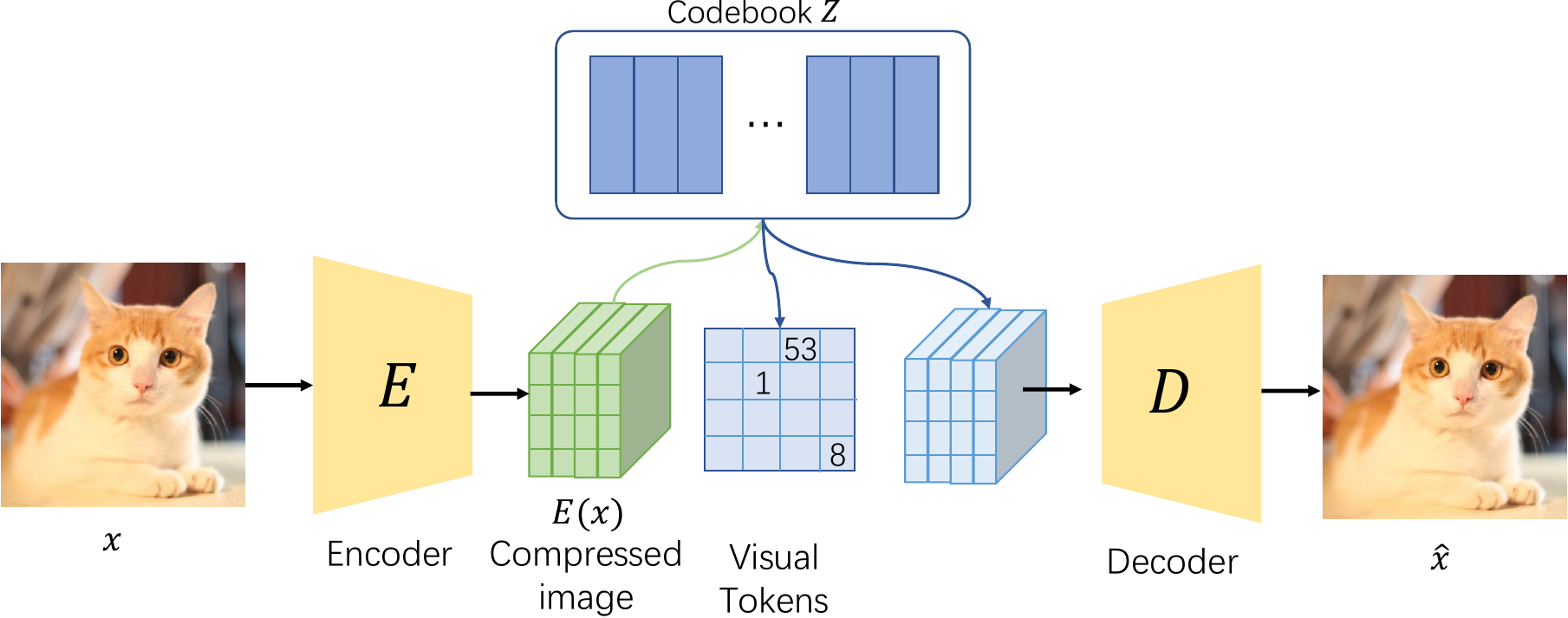}
    \caption{Illustration for the framework of the visual tokenizers.}
    \label{fig:VQGAN}
\end{figure}

Inspired by language models where each word is tokenized by a discrete tokenizer, a series of works also transform images into discrete tokens. Typical visual tokenizers include the VQ-VAEs~\cite{VQVAE1,VQVAE2} and VQGANs~\cite{VQGAN1,VQGAN2}, whose overall framework is shown in Fig.~\ref{fig:VQGAN}. We will begin our discussion with VQ-VAE. Basically, VQ-VAE works as an auto-encoder with an encoder $E(\cdot)$ and a decoder $D(\cdot)$. Given an image $x$, VQ-VAE first encodes it with an encoder $E(\cdot)$ into a lower-dimensional continuous vector $E(x)$. Then, the continuous vector is discretized using a codebook $Z=\{z_k\}_{k=1}^{K}$. The codebook functions similarly to a word embedding table in NLP, where $K$ corresponds to the vocabulary size, and each $z_k \in \mathbb{R}^{n_c}$ represents a visual prototype analogous to a word embedding. With the encoded vector $E(x)$ and the codebook $Z$, we obtain a discrete representation $z_q$ of the image by finding the nearest neighbor of $E(x)$ in $Z$ and use it to reconstruct the image with the decoder: $\hat{x} = D(z_q)$. This provides a way to convert between images and discrete tokens. 

Compared to VQ-VAEs, VQGAN~\cite{VQGAN1, VQGAN2} utilizes a GAN perceptual loss to replace the L2 reconstruction loss, which helps to learn a rich codebook. We use a simple example to illustrate the tokenization process. If we have an input image of size $H\times W \times 3$, after the encoder $E$, we obtain a lower-dimension vector $E(x)$ of size $h\times w \times n_c$, where $h<H$, $w<W$, and $n_c$ denote the dimensions of the code. This means that we can obtain $h \times w$ vectors of dimension $n_c$, and for each vector we will find its nearest neighbor in the code book for discretization so that we will finally obtain a discrete sequence of length $h \times w$ to represent the image.

\noindent \textbf{Remark.} On the one hand, VQGAN and VQ-VAE can be used as visual tokenizers to transform an image into discrete tokens, which enables it to be received by LLMs for visual understanding. On the other hand, they can be used to compress an image into a lower-dimensional space, which motivates the well-known latent diffusion model (LDM)~\cite{stablediffusion}.

\subsection{Multi-modal LLM Architectures}
\begin{figure}
    \centering
    \includegraphics[width=\linewidth]{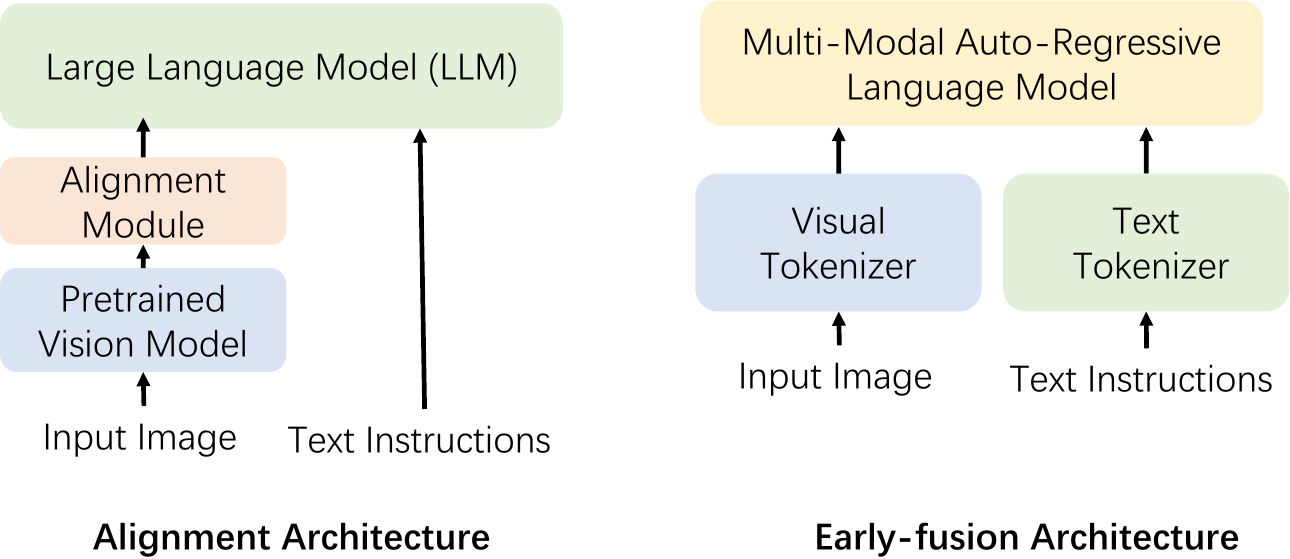}
    \caption{Two branches of multi-modal LLM architectures, including (i) the alignment architecture by aligning pretraining vision models with LLM and (ii) the early-fusion architecture which receives mixed visual and text tokens and relies on autoregressive modeling for multi-modal understanding. }
    \label{fig:MLLM}
\end{figure}

We categorize existing multi-modal LLM architectures into two branches, the alignment architectures and the early-fusion architectures, as shown in Fig.~\ref{fig:MLLM}. Most existing works~\cite{llava, LLaVA-OneVision, huang2024vtimellm} adopt the alignment architecture, which aims to align the vision model from the vision-language pretraining with the pretrained LLM. This branch of models relies on the vision-language pretraining to understand the visual input. After obtaining the embedding of the image, an alignment module such as a projector~\cite{llava} or Q-Former~\cite{blip2} is used to align the image embedding with the LLM space. To train the alignment module, some text-image or text-video pairs are required to input the model. A typical way to align is to make the LLM output the caption of an image given an image embedding. In contrast, as shown on the right of Fig.~\ref{fig:MLLM}, the early-fusion architectures~\cite{chameleon,team2023gemini} do not rely on a pretrained vision model to obtain the semantics of the input image. Instead, similar to NLP, where each word is mapped to a token, the early-fusion architecture maps each visual input into visual tokens through a visual tokenizer. Then, a multi-modal autoregressive language model will receive the mixed text and visual tokens and output the user's desired answers. 

Next, with the overall architecture in mind, we will introduce recent advances in image LLMs and video LLMs.

\subsection{Image LLM} 

We will follow the multi-modal LLM architectures section and elaborate on the latest advancement of image LLM.

\subsubsection{Alignment-Architecture Image LLM}
This architecture treats the image input as an additional extension. The vision encoders are usually frozen and the alignment modules and LLM are tuned based on various strategies to align the multi-modal content and instructions.

\textbf{a) Vision Encoder} is a module that extracts crucial information from images.
Common generic vision encoders include ResNet~\cite{brock2021high}, the CLIP-ViT encoder~\cite{CLIP}, and ImageBind~\cite{girdhar2023imagebind}.
ResNet and CLIP are pretrained on image-text modalities, while ImageBind aligns embeddings from six modalities into one shared space, enabling vision encoders to capture richer information.

\textbf{b) Alignment Module}, also named projector, adapter, etc., aims to mitigate the gap between image features and lexical word tokens and further fuse two modalities. LLaVA~\cite{llava} adopts a simple but effective linear projection to convert image features into word token embedding space and then it concatenates image tokens and word tokens. Such alignment only involves image transformation, limiting interaction with texts, and is not flexible in the visual token number.
Resampler~\cite{alayrac2022flamingo} technique maps varying-size features to a fixed number of tokens.
BLIP-2~\cite{blip2} and MiniGPT-4~\cite{zhu2023minigpt} employ Q-former~\cite{blip2} before linear projections to reduce tokens. Q-former incorporates text semantics and models the interaction between image features and text inputs with learnable queries to enhance the most useful visual content for LLM.
Some works focus on preserving locality during projection, such as Honeybee~\cite{cha2024honeybee}, which introduces a locality-enhanced projector to maintain spatial structure. Others prioritize efficiency, such as TokenPacker~\cite{li2024tokenpacker}, which adopts a coarse-to-fine strategy to compress visual tokens while retaining important details.

\subsubsection{Early-fusion Architecture Image LLM} 
The alignment architecture utilizes the power of off-the-shelf LLM and requires lower computations, but pretrained vision encoders would have information loss and be infected by inductive biases because of the gap between limited pretraining tasks and real demands for image LLM, such as supporting flexible resolution. Therefore, as shown in Fig.~\ref{fig:MLLM}, another line of work aims to train a multi-modal LLM from scratch, where both images and text words are converted into a series of tokens.

Pioneer work Fuyu~\cite{adept_fuyu_8b} adopts linear projections on image patches in spatial order and trains a transformer decoder taking the visual and word token sequence as input. Despite limited performance, it reveals a new technical fashion.
Google follows this fashion, whose Gemini~\cite{team2023gemini} processes the interleaved image and other modalities from the beginning.
Chameleon~\cite{chameleon} trains an image tokenizer that encodes a 512x512 image into 1024 discrete tokens from a codebook of size 8192.
Early-fusion Architecture requires more computation and is more difficult to converge, leaving challenges for future exploration.

\subsubsection{Challenges in Image LLM}
\textbf{(i) Fine-grained visual concept understanding}, where more tokens help encode more detailed information at the cost of causing redundant computation. Chat-UniVi~\cite{jin2024chat} proposes dynamic visual tokens to allocate more computations on important details. An important part of fine-grained understanding is the spatial awareness of object concepts.
AnyRef~\cite{he2024multi} applies RoIAlign to encode regions and designs a segment encoder-decoder to learn segmentation from the image LLM's token outputs, which is similar to OMG-LLaVA~\cite{zhang2024omg}, who generates pixel- and object-centric visual tokens before projections and decodes segmentation tokens from LLM's output by OMG-Seg. Different from segmentation supervision, VisionLLM~\cite{wang2024visionllm} and Virtron~\cite{fei2024vitron} use text supervision such as bounding and polygon descriptions by flexible instruction tuning. Fine granularity modeling offers some explanations for LLM. \textbf{(ii) Hallucination} involves errors in objects, attributes, and relations in the forms of judgment or description~\cite{liu2024survey}. Some works~\cite{you2023ferret} try to reduce biases in training data, while some mitigate hallucination by improving model characteristics such as vision encoders~\cite{chen2024internvl} or fusion mechanisms~\cite{jiang2024hallucination}. Human feedbacks~\cite{stiennon2020learning} also play an important role in reducing hallucination.

\noindent \textbf{Remark.} Currently, the alignment architecture still outperforms the early-fusion architecture in multi-modal understanding, e.g., with comparable parameters, the early-fusion architecture Emu3~\cite{wang2024emu3} achieves 75.1 score on VQAv2~\cite{goyal2017making} benchmark and 58.5 score on MMBench~\cite{liu2024mmbench} benchmark, while the early-fusion architecture LLAVA-1.6 achieves 86.8 and 67.4 score, respectively. The advantages and disadvantages of the two architectures are as follows: (i) The advantage lies in the capability of utilizing the pretrained knowledge from the vision encoder and LLM. The vision-language pretraining enables the output of the vision encoder to contain semantic meanings. Only the alignment module needs to be trained, which makes this paradigm resource-friendly. (Sometimes other modules are also learnable for better performance.) However, its ability is also limited by the pretrained vision encoder and LLM, e.g., the pretrained CLIP vision encoder often struggles with multiple objects, making the multi-modal LLMs based on CLIP inherit the limitation. (ii) The disadvantage comes from the fact that the early-fusion architecture may have a higher potential, because all its parameters are trained from scratch. However, training from scratch makes the early-fusion architecture face two challenges: (a) a good visual tokenizer needs to be trained, and (b) more resources will be needed to train the multi-modal autoregressive model. First, since the visual tokenization process involves compression and discretization, there inevitably exists visual information loss. How to train a tokenizer that contains rich visual information still remains a challenging problem. Second, the visual tokenizers are generally trained with the image reconstruction objective, which in essence belongs to a pixel-level task instead of a semantic-level task. This training strategy requires that the downstream multi-modal LLMs should have an additional ability to learn semantic meanings from the pixel-level information, compared to the original LLMs, which are only expected to understand semantic tokens. Therefore, multi-modal LLMs tend to require more data for training. 

\subsection{Video LLM} 

Following the success of Image LLMs, researchers start exploring the training of Video LLMs~\cite{vidllmsurvey}. Typically, videos are viewed as sequences of image frames (some Video LLMs incorporate other modalities like audio or speech), so Video LLMs have a higher computational complexity. The challenge of collecting high-quality video datasets further complicates the training process, making early fusion architectures computationally exhaustive. As a result, almost all the existing Video LLMs adopt the alignment architectures.

\subsubsection{Alignment-Architecture Video LLM}
The video LLM architecture is similar to that of Image LLMs with alignment architectures. By sampling a fixed number of frames or using a fixed frames-per-second (FPS) rate, videos are reduced to a limited set of images. The visual embeddings of each image are then extracted using a visual encoder. These features are sequentially concatenated in the order of the frames and connected to the LLM via an alignment module. In earlier works, VideoChat~\cite{2023videochat} utilizes a Q-former structure as the alignment module, while VideoLLaMA~\cite{zhang2023video} introduces an audio encoder and an audio Q-former to handle audio signals. Video-ChatGPT~\cite{Maaz2023VideoChatGPT} takes a different approach by average-pooling each frame's patch embeddings along the spatial and temporal dimensions before using a linear layer as the alignment module. Training Video LLMs also follow an ``alignment then instruction tuning'' strategy. While additional GPT-annotated or human-annotated video datasets are collected, image datasets can also be leveraged by treating images as single-frame videos.

Recent successful efforts focus on improving performance by refining the alignment module and scaling up the model and dataset sizes. For instance, VideoLLaMA2~\cite{damonlpsg2024videollama2} improves the alignment module to model the connections across temporal and spatial dimensions. It also gathers datasets for tasks such as captioning, classification, and question answering. Qwen2.5-VL~\cite{bai2025qwen25vl} and InternVL3~\cite{zhu2025internvl3} leverage diverse training data, including images, videos, and interleaved image–text pairs, to build powerful vision-language models.

\subsubsection{Challenges and Limitations in Video LLM}
Compared to Image LLMs, Video LLMs face two unique challenges. The first challenge is understanding videos at a finer granularity, specifically the comprehension of video segments and the relationships between these segments. The second challenge is understanding long-form videos, such as movies, within the limited context length of LLMs.

For segment-level video understanding, VTimeLLM~\cite{huang2024vtimellm} transforms the temporal video grounding and dense video captioning tasks into a sequence-to-sequence format. After alignment training, it introduces an additional boundary perception training, leveraging large-scale multi-event video-text data to enhance awareness of event boundaries and timestamps. Finally, it incorporates temporal reasoning data during instruction tuning. Some approaches~\cite{Grounding-Prompter, feng2023llm4vg} adopt training-free methods, where sampled frames are individually captioned, and each frame's timestamp and caption are input into an LLM via carefully crafted prompts, allowing the LLM's powerful reasoning capabilities to comprehend each segment.

For long-form videos, traditional Video LLMs struggle with input limitations. For example, a Q-former in BLIP-2 encodes an image into 32 tokens; sampling 256 frames results in 8K tokens, which reaches the maximum context length of most LLMs. However, this represents less than 5 minutes of video at a sampling rate of 1 FPS. Therefore, more efficient representations are necessary for processing long-form videos like movies. MovieChat~\cite{song2024moviechat} introduces a memory consolidation mechanism that merges similar image tokens once the token limit is reached. LWM~\cite{LWM} and LongVA~\cite{LongVA} handle long video inputs by using LLMs with larger context lengths and more efficient attention mechanisms. Some methods~\cite{huang2024vtimellm, LLaMA-VID} reduce the number of tokens per frame, representing each frame with only 1 or 2 tokens on average. Other approaches~\cite{wang2024videotree} convert long-form videos into text corpus using image captioning and employ LLMs as agents to search for specific answers within the text corpus.

\noindent \textbf{Remark.} Despite the advancements in Video LLMs, nearly all existing models rely on sampling frames and encoding them individually through image encoders. This approach may be favored due to several reasons: image encoders are less computationally intensive compared to video encoders, they offer better alignment with textual data, and they facilitate unification with Image LLMs. However, this methodology comes with a significant limitation. Specifically, the process of sampling frames can lead to the complete loss of information that occurs between sampled frames. As a result, these models fail to capture the continuous motion and trajectories of objects, which are essential for understanding dynamic scenes and activities within a video.

\subsection{Speech LLM}
 
Similar to Image LLMs, the architecture of Speech LLMs can generally be categorized into two types: alignment-based architectures and early-fusion architectures~\cite{peng2025survey}.

\subsubsection{Alignment-Architecture Speech LLM}
This architecture first extracts information from audio with pre-trained or fine-tuned audio encoder and produces audio embedding.

\textbf{a) Audio Encoder} transforms raw waveforms into time–frequency representations using conventional signal processing techniques. The most commonly used audio encoders are Whisper~\cite{radford2023robust} and Conformer~\cite{gulati2020conformer}. Whisper is an automatic speech recognition (ASR) model with an encoder–decoder Transformer architecture, similar to sequence-to-sequence models in natural language processing. It is trained on 680,000 hours of multilingual, multitask supervised data collected from the web, covering speech recognition, speech translation, and language identification. Conformer (Convolution-augmented Transformer) combines convolutional neural networks (CNNs) with Transformer blocks, effectively capturing both local and global dependencies in speech signals. Other widely adopted encoders include WavLM~\cite{chen2022wavlm}, a self-supervised speech representation model built on the HuBERT~\cite{hsu2021hubert} framework, with improvements in pretraining objectives and data diversity.

\textbf{b) Alignment Module} also referred to as a projector, connector, or adapter, maps audio embeddings into the text embedding space, enabling them to be processed by the LLM decoder for downstream understanding tasks. Several types of alignment modules have been proposed. One common approach is a multi-layer perceptron (MLP), which performs a straightforward projection. Another is the Q-Former, which introduces trainable query tokens that attend to audio features and produce fixed-length embeddings compatible with the LLM input space. A third approach is cross-attention, which allows bidirectional interactions between audio and text features, facilitating richer multimodal integration.

\subsubsection{Early-fusion Architecture Speech LLM}
This type of Speech LLMs is inspired by visual tokenizers and adopts a similar approach for audio. In this framework, raw audio is converted into a sequence of discrete tokens that capture the acoustic content and can often be decoded back into high-quality audio. The generation of discrete tokens relies on vector quantization (VQ). Building on VQ-VAE~\cite{van2017vqvae}, which introduced the idea of encoding continuous audio features into symbolic representations via a learned codebook, modern approaches include self-supervised pre-trained audio tokenizers such as HuBERT~\cite{hsu2021hubert} and neural codec models such as EnCodec~\cite{defossez2022high}. Several representative works fall under this branch of Speech LLMs. VALL-E~\cite{chen2024vall} leverages EnCodec tokens to achieve zero-shot speech synthesis. SpeechGPT~\cite{zhang2023speechgpt} is trained on paired unit-text data, where spoken audio is represented as discrete speech units. AudioPaLM~\cite{rubenstein2023audiopalm} integrates wav2vec-style audio tokenization with language modeling to improve multimodal speech understanding.

Now we have discussed the multi-modal LLM for understanding. Next, we will discuss another important topic of multi-modal generative AI, i.e., multi-modal diffusion models for generation.   

\section{Multi-modal Diffusion for Generation}
Diffusion models have been one of the most successful generative models in visual generation given texts and are widely used in multi-modal generation tasks. We present the famous latent diffusion model~\cite{stablediffusion}, and discuss several advanced diffusion-based text-to-image and text-to-video models.

\begin{figure}
    \centering
    \includegraphics[width=\columnwidth]{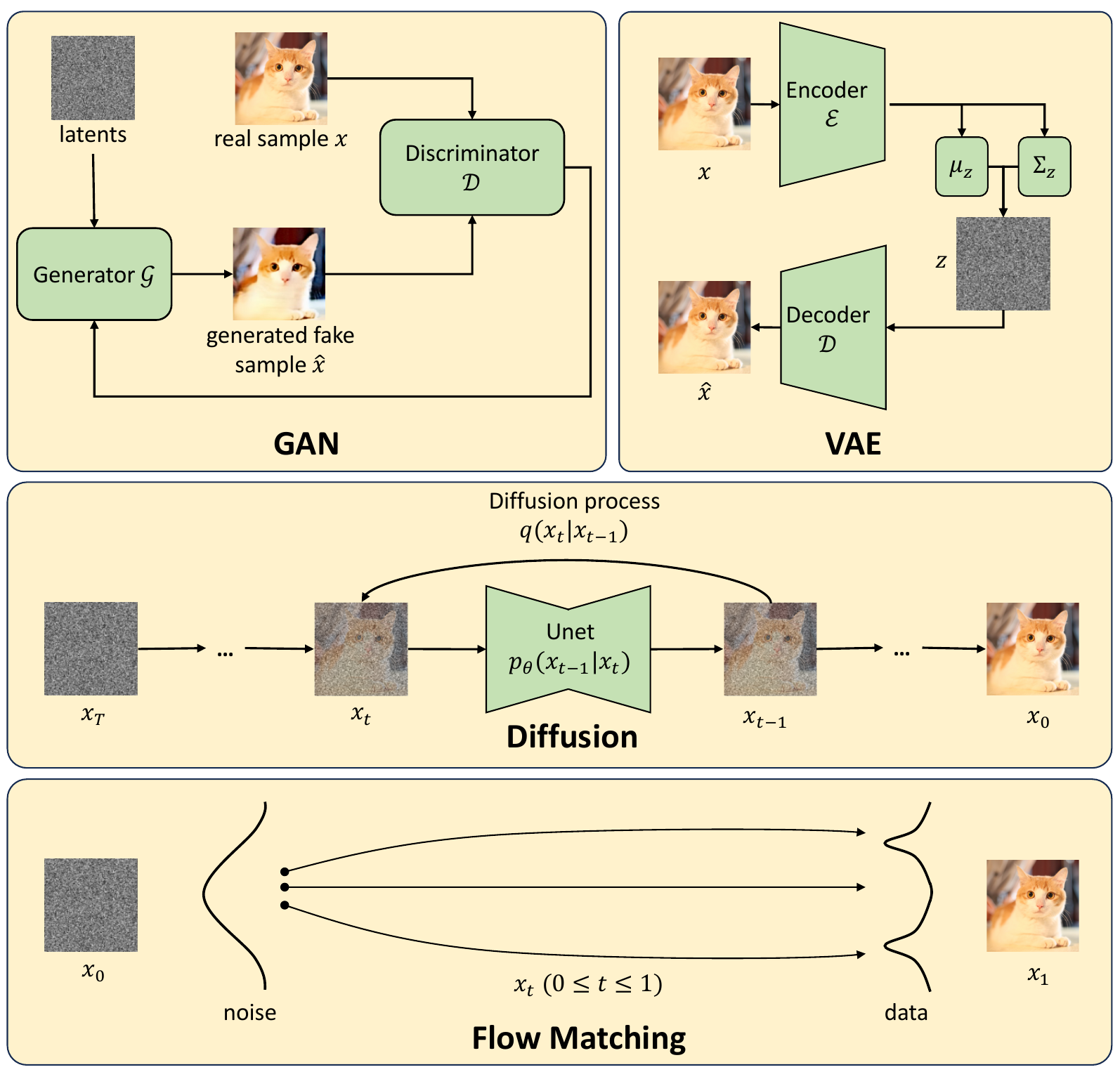}
    \caption{\zyw{Comparison among GAN, VAE, diffusion, and flow matching models.}}
    \label{fig:gen-cmp}
\end{figure}

\label{sec:diffusion}
\subsection{Preliminaries}
\label{subsec:diff:preliminary}
We will first introduce some preliminaries, including traditional generative models, i.e., generative adversarial networks (GANs) and Variational AutoEncoders (VAEs). 
\zyw{We then introduce diffusion probabilistic modeling and present a comparison among GAN, VAE, diffusion, and flow matching models, as illustrated in Fig.~\ref{fig:gen-cmp}.}
 
\subsubsection{Generative Adversarial Networks}
\label{subsubsec:gan:preliminary}

The generative adversarial network (GAN)~\cite{goodfellow2014gan} is one of the earliest neural architectures designed to generate visual content such as images~\cite{bao2017cvae} and videos~\cite{vondrick2016generating}. The main idea of GANs involves two networks: a generator $\mathcal{G}$ and a discriminator $\mathcal{D}$. Specifically, $\mathcal{G}$ aims to generate visual content from a noise vector $z$, while $\mathcal{D}$ is trained to distinguish between real visual samples $x$ and generated ones $\mathcal{G}(z)$. These two networks are trained in an adversarial manner: the generator tries to produce outputs that can fool the discriminator, and the discriminator strives to accurately classify real versus fake samples. The training process forms a min-max game, where the generator learns to generate increasingly realistic samples to deceive a progressively stronger discriminator. The two networks are mutually reinforcing, so the training objective is as follows:
\begin{equation}
    \underset{\mathcal{G}}{min}\; \underset{\mathcal{D}}{max}\; \mathbb{E}_{x\sim p_x}\log \mathcal{D}(x) + \mathbb{E}_{z\sim p_z}\log(1-\mathcal{D}(\mathcal{G}(z))),
\end{equation}
where $z$ is sampled from $p_z$ that is usually a normal distribution and $x$ is a sample from the real data distribution $p_x$.

\subsubsection{Variational AutoEncoder} Variational AutoEncoder~\cite{kingma2014vae} (VAE) is another typical generative model. Unlike GANs, autoencoders have an encoder-decoder architecture that uses an encoder $\mathcal{E}$ to present the visual content $x$ to a latent code $z = \mathcal{E}(x)$ and a decoder $\mathcal{D}$ to reconstruct the data $\hat x = \mathcal{D}(z) \approx x$. However, normal autoencoders have no constraints on the latent space, which makes them overfit the dataset easily. To solve the problem, VAEs make a regularization to the latent space and sample $z$ from a distribution $p_\theta$, typically a Gaussian distribution, where $\theta$ is the parameters of the encoder-decoder model.
As the distribution $p_\theta$ is unknown, VAE utilizes a recognition model $\phi$ which serves as a variational approximation $q_\phi$ to approximate $p_\theta$ and trains them jointly:
\begin{equation}
    \mathcal{L}(\theta, \phi;x) = -D_{KL}(q_\phi(z|x) || p_\theta(z)) + \mathbb{E}_{q_\phi(z|x)}[\log p_\theta (x|z)],
\end{equation}
where $D_{KL}$ means the Kullback-Leibler divergence. $\phi$ can be formulated as a differentiable estimator using the parameterization trick. To better generate visual content, many efforts~\cite{mittal2017sync,van2017vqvae,li2018videofromtext} have been made based on VAE. Sync-DRAW~\cite{mittal2017sync} introduces a novel architecture that combines VAE with a recurrent attention mechanism to create a unique temporally dependent sequence of frames. 

Despite the successful introduction of VAEs, they still face a significant issue where the model ignores the information in the latent space and relies solely on a powerful decoder to reconstruct the data, a phenomenon known as ``posterior collapse''. To address this problem, the VQ-VAE~\cite{van2017vqvae} utilizes discrete encoding to learn the prior and employs vector quantization methods to prevent the latents from becoming uninformative.

\subsubsection{Diffusion Probabilistic Modeling}
\label{subsubsec:diff:prob}

Compared to GANs and VAEs, a new branch of generative models, diffusion models~\cite{ddpm,ddim,stablediffusion} have become dominant in many tasks such as text-to-image generation or text-to-video generation. The core idea of diffusion modeling is to learn the transformation between the real data distribution $q(x_0)$ and a standard Gaussian distribution $q(x_T)$. 

We briefly introduce the denoising diffusion probabilistic model (DDPM), which includes the forward and backward processes. In the forward process, given a real data sample $x_0$, it will go through a Markov process with more and more random Gaussian noise added to the sample as follows:
\begin{equation}
\label{eq:ddpm_f}
q(x_t|x_{t-1}) = \mathcal{N}(x_t; \sqrt{1-\beta_t}x_{t-1}, \beta_tI) , t=0,1,\cdots, T
\end{equation}
where $t$ is the time step, $T$ is usually large so that $x_T$ is close to a Gaussian noise, and $\beta_t$ is a parameter to control the noise schedule. Conversely, to achieve generation from random noise, what DDPM does in the backward process is to learn the following distribution:
\begin{equation}
\label{eq:ddpm_b}
p_{\theta}(x_{t-1}|x_t) = \mathcal{N}(x_{t-1}; \mu_{\theta}(x_t,t), \Sigma_{\theta}(x_t,t)),
\end{equation}
where a neural network parameterized by $\theta$ is designed to predict the less noisy image $x_{t-1}$. Then, with this denoising network $\theta$, we can denoise from a random noise $x_T$ step by step until we get a clean data sample $x_0$, which could be an image or a video, etc.

\noindent \textbf{Remark.} GANs, VAEs, and diffusion models are all generative models. Compared to GANs, which train both the generator and discriminator, the diffusion models have explicit probabilistic modeling and only train a denoising network $\epsilon_{\theta}$, which is more stable. Similarly, VAEs train both an encoder and a decoder. Moreover, diffusions denoise for each image $T$ times in the training phase, resulting in $T$ variants of each image as augmentation. These augmented images in turn help the denoising network to better model the data distribution $p_{\theta}(x_0)$, leading to better generation results.

\subsubsection{Latent Diffusion Model}
\begin{figure}
    \centering
    \includegraphics[width=\linewidth]{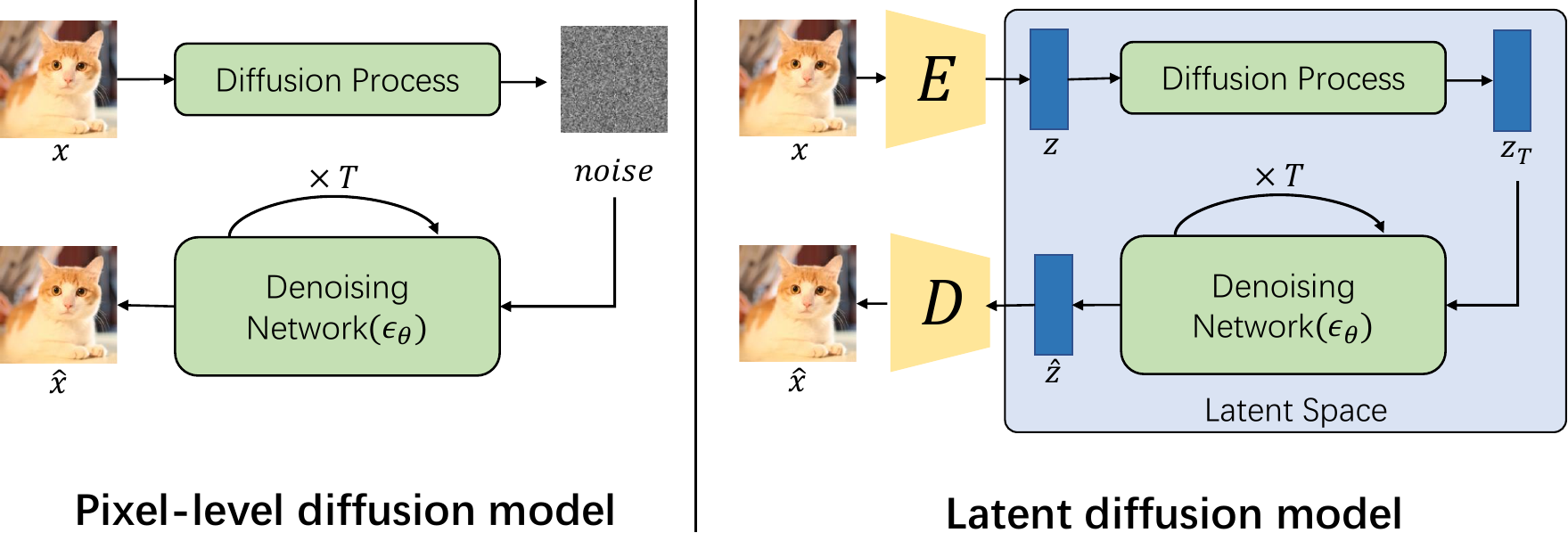}
    \caption{Comparison between pixel-level and latent diffusion models.}
    \label{fig:LDM}
\end{figure}

As shown in Eq.~\eqref{eq:ddpm_f} and Eq.~\eqref{eq:ddpm_b}, the denoising process of diffusion models is conducted on the pixels of each image in an iterative manner, which results in high computational cost, especially when the generated image is high-resolution. To tackle this problem, the latent diffusion model (LDM)~\cite{stablediffusion} proposed to conduct the diffusion process in the latent space instead of the pixel space. The framework comparison between the pixel-level diffusion model and LDM is shown in Fig.~\ref{fig:LDM}. To reduce the computational cost, LDM utilizes the encoder of VQGAN~\cite{VQGAN1} to compress the image into the latent space, $z=E(x)$, which has a much lower dimension than the original image. Then, the diffusion process in Eq.~\eqref{eq:ddpm_f} and Eq.~\eqref{eq:ddpm_b} will be conducted in the latent space.

Note that there is an additional input $c$ of the denoising network that is for conditional generation, e.g., as for the text-to-image generation task, $c$ could be the representation of the text prompt~\cite{reed2016generative}. 
Also, $c$ could be other conditions, such as layout~\cite{he2023localized} or semantic maps~\cite{isola2017image}. Since most computation, including the training and iterative inference, is conducted in the lower-dimension latent space, the LDM model exhibits high efficiency. Therefore, most text-to-image and text-to-video models adopt the LDM structure.

\subsubsection{Flow Matching}
Compared with diffusion models such as DDPM, Flow Matching~\cite{lipmanflow} represents a new paradigm in generative modeling, built upon Continuous Normalizing Flows (CNFs). It introduces a simple yet intuitive training objective that learns to approximate a target vector field, which defines a probability path transforming noise samples into data samples. In this way, diffusion processes can be viewed as special cases within the broader Flow Matching framework.

Let $x_1$ denote a random variable drawn from an unknown data distribution $q(x_1)$. We define a probability path $p_t$ such that $p_0=p$ is a simple distribution, e.g., the standard normal distribution $p(x)=\mathcal{N}(x|0,I)$, and $p_1$ approximates the data distribution $q$. The goal of Flow Matching is to learn a vector field that aligns the model’s probability path with this target path from $p_0$ to $p_1$.
\begin{equation}
  \mathcal{L}_{\rm FM}(\theta)=\mathbb{E}_{t,p_t(x)}\Vert v_t(x)-u_t(x)\Vert^2,
\end{equation}
where $p_t(x)$ denotes the target probability density path, $u_t(x)$ is the corresponding vector field, and $v_t(x,\theta)$ is the learnable CNF vector field parameterized by $\theta$. Here $t\sim\mathcal{U}[0,1]$ is the uniform distribution, and $x\sim p_t(x)$. In essence, the Flow Matching loss trains the neural vector field $v_t$ to regress toward the target field $u_t$. When the loss approaches zero, the learned CNF model successfully reproduces the probability path $p_t(x)$.

\subsection{Text-to-Image Generation}
\label{sec:t2i generation}

As mentioned in the preliminary part, diffusion models can be broadly categorized into two branches: pixel-based and latent-based~\cite{zhang2023text}. In the early development stage, the denoising process is typically applied directly in the pixel space. For instance, GLIDE~\cite{nichol2021glide} is a pioneering work in photorealistic image generation with text guidance, using a 3.5 billion parameter diffusion model that employs a text encoder to condition on natural language descriptions. GLIDE also explores the use of CLIP guidance and classifier-free guidance in diffusion models, finding that classifier-free guidance produces higher-quality images. Besides, Imagen~\cite{saharia2022photorealistic} follows GLIDE and adopts classifier-free guidance for its pixel-based diffusion model. The key difference between them is that GLIDE trains a text encoder and a diffusion model together with text-image pairs, while Imagen utilizes pretrained and frozen large transformer language models, leveraging their strong text understanding capabilities to enhance sample fidelity and image-text alignment.

However, directly operating in pixel space requires substantial computational resources, which leads to the appearance of latent-based diffusion models. A milestone in this area is Stable Diffusion~\cite{stablediffusion}, which introduces the concept of latent diffusion model to strike a near-optimal balance between complexity reduction and detail preservation. It incorporates a pretrained VQGAN to compress images from pixel space into semantic latent space. Compared to pixel-based diffusion methods, Stable Diffusion not only achieves competitive performance across multiple image generation tasks but also significantly reduces both training and inference costs. Another notable example of a latent-based model is DALL-E2~\cite{ramesh2022hierarchical}, which combines a CLIP model and a diffusion model to enable zero-shot text-guided image generation. DALL-E2 consists of a CLIP image encoder and a diffusion decoder that inverts the encoder, allowing for explicit generation of image representations. This approach improves image diversity while maintaining photorealism and caption similarity.

GLIDE~\cite{nichol2021glide}, Imagen~\cite{saharia2022photorealistic}, Stable Diffusion~\cite{stablediffusion}, and DALL-E2~\cite{ramesh2022hierarchical} are all pioneering works that represent different technological pathways in the field of text-to-image generation. These models have greatly inspired subsequent research and development~\cite{chen2023disenbooth,chen2023videodreamer,chen2024disenstudio}. Despite their differences, some common trends have emerged in their development. First, latent-based diffusion methods have become increasingly prevalent due to their advantages in conserving computational resources and generating high-quality images. Second, compared to classifier guidance~\cite{dhariwal2021diffusion}, classifier-free guidance~\cite{ho2022classifier} is widely adopted in these works, where the label in a class-conditional diffusion model is replaced with a null label at a fixed probability during training. Third, U-Net traditionally serves as the backbone of the diffusion model, facilitating denoising and the gradual generation of high-quality images.

\begin{figure}
    \centering
    \includegraphics[width=\linewidth]{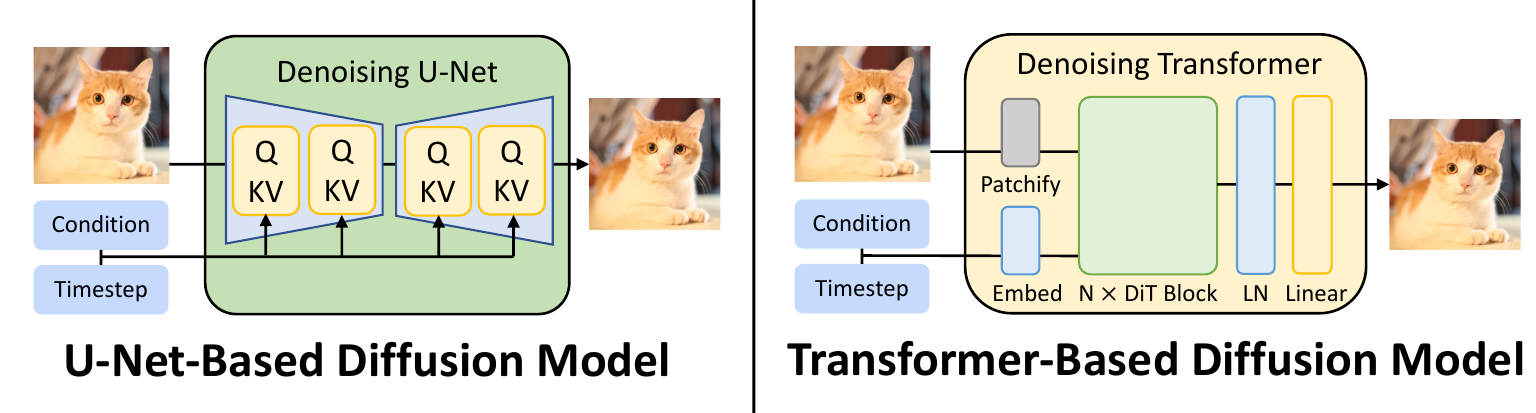}
    \caption{Comparison between U-Net-based diffusion model and Transformer-based diffusion model.}
    \label{fig:DiT}
\end{figure}

Despite its advantages in high-resolution image generation, U-Net's specific structures, such as ResBlocks and convolutional operations, limit its scalability. In contrast, Transformers, which are better suited to handle larger-scale data and tasks, are emerging as strong contenders to U-Net. The Diffusion Transformer (DiT)~\cite{peebles2023scalable} represents a class of diffusion models that replaces the commonly used U-Net backbone with a transformer backbone, as shown in Fig.~\ref{fig:DiT}. This approach is supported by empirical findings suggesting that the U-Net inductive bias is not crucial to the performance of diffusion models. Additionally, utilizing a transformer backbone enables the diffusion model to leverage the best practices of transformers, such as architectural design and training paradigms, along with their good properties, such as scalability, robustness, and efficiency. Specifically, DiT adheres to the foundation of the Latent Diffusion Model (LDM) framework and emulates the design of the Vision Transformer (ViT) by introducing a comprehensive DiT design space, including patch size, transformer block architecture, and model size. The first layer of DiT, termed patchify, converts the spatial input into a sequence of tokens by linearly embedding each patch. Following the patchify step, the input tokens are processed through a sequence of transformer blocks that incorporate conditioning, such as time and label. The proposed transformer design includes adaptive layer norm (adaLN) block, cross-attention block, and in-context conditioning block. After the final block, a transformer decoder translates the image tokens into output predictions. The difference between U-Net-based and Transformer-based diffusion models is illustrated in Fig.~\ref{fig:DiT}.

\begin{figure}
    \centering
    \includegraphics[width=\linewidth]{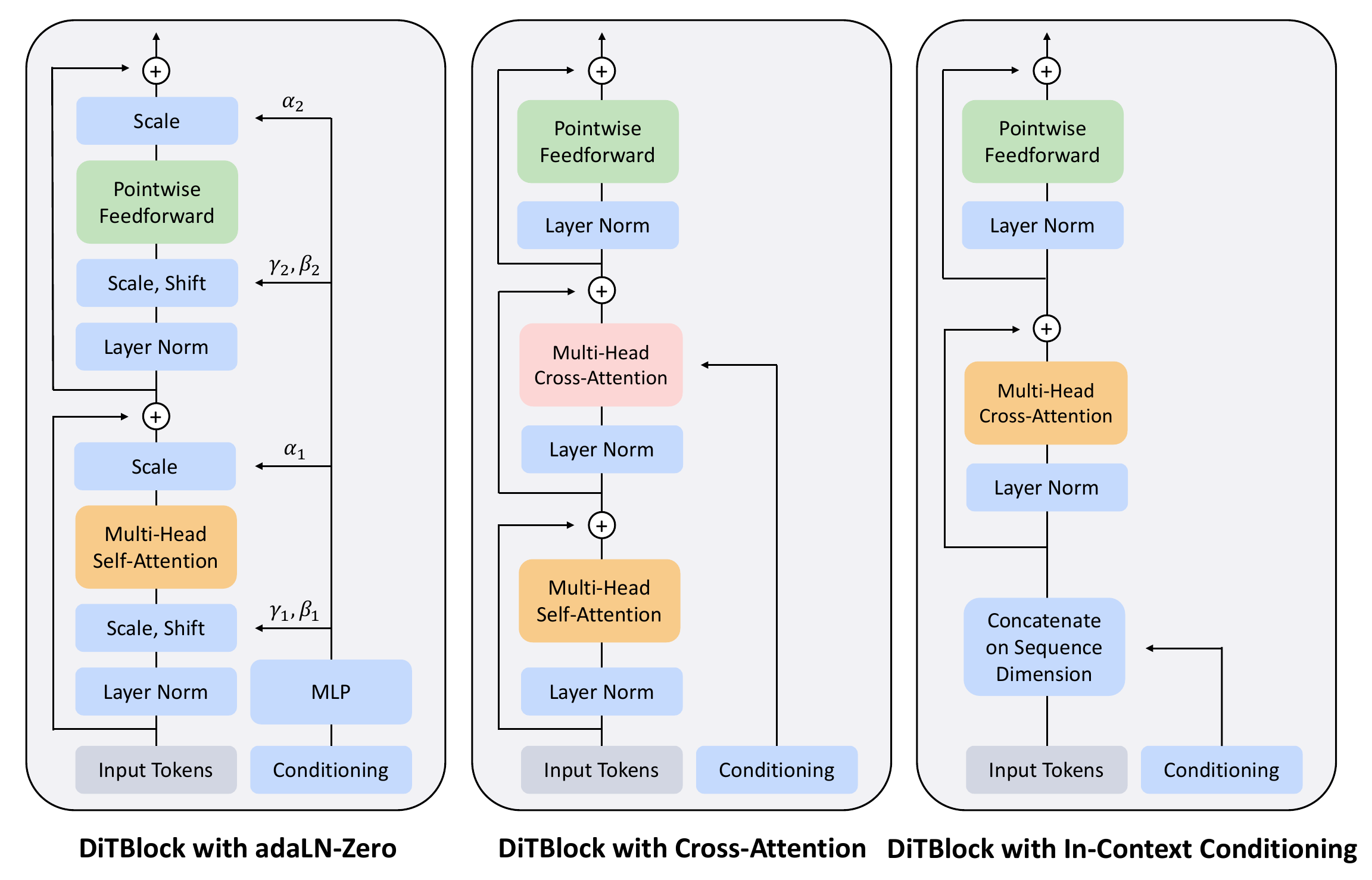}
    \caption{Comparison between different DiT blocks from~\cite{peebles2023scalable}.}
    \label{fig:DiT-Block}
\end{figure}

The three distinct transformer blocks are the core modules of DiT, representing different ways to interact with multi-modal information, including images, timestep, and conditions. Their designs are inspired by the standard ViT block design but incorporate small yet significant modifications. As illustrated in Fig.~\ref{fig:DiT-Block}, these blocks differ in how the image latent interacts with the conditioning information. The adaLN block follows the adaptive normalization layers in GANs, replacing the standard normalization layers in transformer blocks. The scale and shift parameters in this block are determined by the sum of the embedding vectors of timestep and condition. This block adds the least Gflops to the model. The cross-attention block introduces an additional multi-head cross-attention layer, serving as the interaction module between the image latent and the timestep and condition. This block adds the most Gflops to the model. The in-context conditioning block treats the tokens from the timestep and condition in the same way as image tokens, concatenating them along the sequence dimension. This block introduces a moderate amount of Gflops.

Following the development of DiT~\cite{peebles2023scalable}, a growing number of works are exploring variants of diffusion transformers with improved performance. For instance, CrossDiT~\cite{chen2023pixart} combines the adaLN-zero DiT block and cross-attention DiT block. It simplifies adaLN-zero layers to adaLN-single layers by removing label conditioning and using only time conditioning for scale and shift control. It incorporates text embeddings from T5~\cite{raffel2020exploring} into the multi-head cross-attention layer. Another notable variant is MM-DiT~\cite{esser2024scaling}, which integrates the adaLN-zero DiT block and in-context conditioning DiT block. This model uses text embeddings from CLIP and timestamps to condition the network, employs two separate sets of weights for image and condition modalities, and concatenates image and condition for the attention operation. Empirical experiments show that both CrossDiT and MM-DiT outperform the vanilla DiT in terms of validation loss, CLIP score, and FID.

The designs of diffusion transformer variants are distinct from each other, but they basically derive from the three core architectures proposed by DiT: the adaLN-zero block, the cross-attention block, and the in-context conditioning block. Currently, MM-DiT, which combines the adaLN-zero block with in-context conditioning, represents the state-of-the-art architecture. Its advantage lies in training the text modality iteratively alongside the diffusion process in an in-context manner rather than keeping it frozen, which produces a more diverse semantic space.

\subsection{Text-to-Video Generation}
\label{sec:t2v generation}

Due to the success of diffusion models in text-to-image tasks, many researchers have introduced temporal information to the diffusion models and utilized the capability of generating high-quality images to conduct text-to-video models.

The most intuitive approach to utilizing the text-to-image model is modifying the self-attention mechanism, which gets the text-to-video model without any additional parameters. Text2Video-Zero~\cite{khachatryan2023t2vzero} is one of the pioneer works. Rather than randomly initializing the latents of all frames independently, Text2Video-Zero only samples the latent code $z_T^1$ of the first frame and applies $\Delta t$ DDIM backward steps to obtain $z_{T'}^1$. After that, Text2Video-Zero determines the global scene and a camera motion direction, proposes a warping function $W_k$ to get all $F$ frames from $z^1_{T'}$ to $z^F_{T'}$, and then performs a DDPM forward to get the initial latents. To keep the consistency among different frames, Text2Video-Zero proposes cross-frame attention, which uses keys and values from the first frame to generate the images. Latent-Shift~\cite{an2023latentshift} is another representative method. It proposes a novel Temporal-Shift module that splits the latents along the channel dimension and shifts the split channel along the temporal dimension to keep the consistency of all frames. These methods have fully used the powerful pretrained text-to-image models and can generate videos with much higher resolution and quality than traditional text-to-video methods using GANs and VAEs. However, rather than capturing, training, and understanding the temporal information, these methods are more like providing a class of expert knowledge that can utilize the temporal information from a human perspective. Thus, these methods enjoy high generation efficiency, but the videos generated still struggle with motion smoothness and video consistency.

To solve the problems, another kind of approaches~\cite{ho2022videodiffusionmodels,singer2024makeavideo,guo2024animatediff} not only inherits the architecture of the T2I models but also makes efforts to introduce novel modules or modify the original structure to learn the temporal information. VDM~\cite{ho2022videodiffusionmodels} is one of the earliest works that transferred the T2I model to solve T2V tasks. VDM proposes a 3D U-Net that modifies the diffusion architecture by changing each 2D spatial convolutional layer into a 3D convolution. After that, for each spatial attention block, VDM inserts a temporal attention block that performs attention over all frames with relative position embeddings to distinguish the ordering of frames. Make-a-video~\cite{singer2024makeavideo} proposed a pseudo-3D convolutional and attention layer, which consists of a spatial 2D convolutional layer and a temporal 1D convolutional layer. Compared to 3D convolution, this approach is much more efficient while facilitating information sharing between the spatial and temporal axes. To more flexibly apply the capabilities of the T2I model, such as the customization and style transferring ability brought by LoRA, AnimateDiff~\cite{guo2024animatediff} keeps the original architecture and only inserts a motion module after each pretrained layer. The motion module consists of an input projection layer, several temporal self-attention layers, and an output projection layer. To avoid harming the original capabilities of T2I models, AnimateDiff zero initializes the output projection layer. 

As the attention-based architecture is more suitable for capturing long-range contextual relationships, some methods~\cite{ma2024latte,chen2024gentron} adopt a DiT-based model to generate videos. Latte~\cite{ma2024latte} utilizes a video transformer as the backbone and employs a VAE to encode videos into features, which is used to extract tokens. Currently, compared to U-Net-based methods, DiT-based methods can scale to larger datasets and parameters, hence yielding relatively better performance. However, this also implies a higher consumption of computational resources. The DiT-based methods are commonly adopted in accomplishing some outstanding applications within the industry.

\subsection{Text-to-Speech Generation}
Text-to-Speech (TTS) generation, also known as speech synthesis, is one of the most fundamental tasks in multimodal speech processing~\cite{zhang2023survey}. The development of TTS has evolved from a three-stage pipeline to a two-stage framework, and more recently, to end-to-end systems. Before the advent of neural networks, TTS systems typically converted text into linguistic features and then into acoustic features before decoding them into waveforms. With the introduction of neural networks, this process was simplified, where text only needs to be transformed into either linguistic or acoustic representations. Most recent diffusion-based TTS models adopt a two-stage approach: an acoustic model first generates acoustic features, which are then converted into waveforms using a vocoder. Moreover, several studies explore end-to-end TTS frameworks that directly synthesize speech waveforms from text input.

For two-stage text-to-speech diffusion models, the acoustic model and vocoder are the two key components. The acoustic model converts text into acoustic representations, while the vocoder synthesizes waveforms from these features. DiffWave~\cite{kong2020diffwave} is one of the earliest diffusion-based speech synthesis models, serving as a neural vocoder. It formulates waveform generation as a DDPM task, where a neural network learns to reverse a gradual noising process applied to real waveforms. WaveGrad~\cite{chen2020wavegrad} also functions as a vocoder, introducing a continuous-time, score-based diffusion approach that models a gradient field to guide the denoising process, rather than relying on a discrete noise schedule. Grad-TTS~\cite{popov2021grad} is a diffusion-based acoustic model that extends diffusion modeling from vocoders to full TTS systems. It generates acoustic features from text through stochastic differential equations (SDEs), enabling a non-autoregressive acoustic modeling framework. Diff-TTS~\cite{jeong2021diff} is another diffusion-based acoustic model that further advances speech synthesis by formulating the entire acoustic modeling process as a deterministic or stochastic denoising procedure.

Compared with two-stage approaches, end-to-end text-to-speech diffusion models reduce error propagation and produce higher-quality speech, becoming the mainstream development direction. For example, WaveGrad 2~\cite{chen2021wavegrad} discards the two-stage design of WaveGrad~\cite{chen2020wavegrad} and adopts an end-to-end framework that directly generates audio from a phoneme sequence. Moreover, recent systems such as TTS-1~\cite{atamanenko2025tts} and MiniMax-Speech~\cite{zhang2025minimax} also follow end-to-end architectures and achieve remarkable performance in speech generation.

\section{Unification of Understanding and Generation}
\label{sec:unified framework}
Until now, we have discussed both the multi-modal LLMs and the multi-modal diffusion models, where the former works well for multi-modal understanding and the latter exhibits a powerful ability in visual generation. Then a natural question arises: could we have a unified model that can simultaneously work well for multi-modal understanding and generation? Next, we will discuss this trending problem from the following two perspectives: (i) the probabilistic modeling method, and (ii) the model architecture. 

\subsection{Probabilistic Modeling: Autoregressive or Diffusion?}
\begin{figure}
    \centering
    \includegraphics[width=\linewidth]{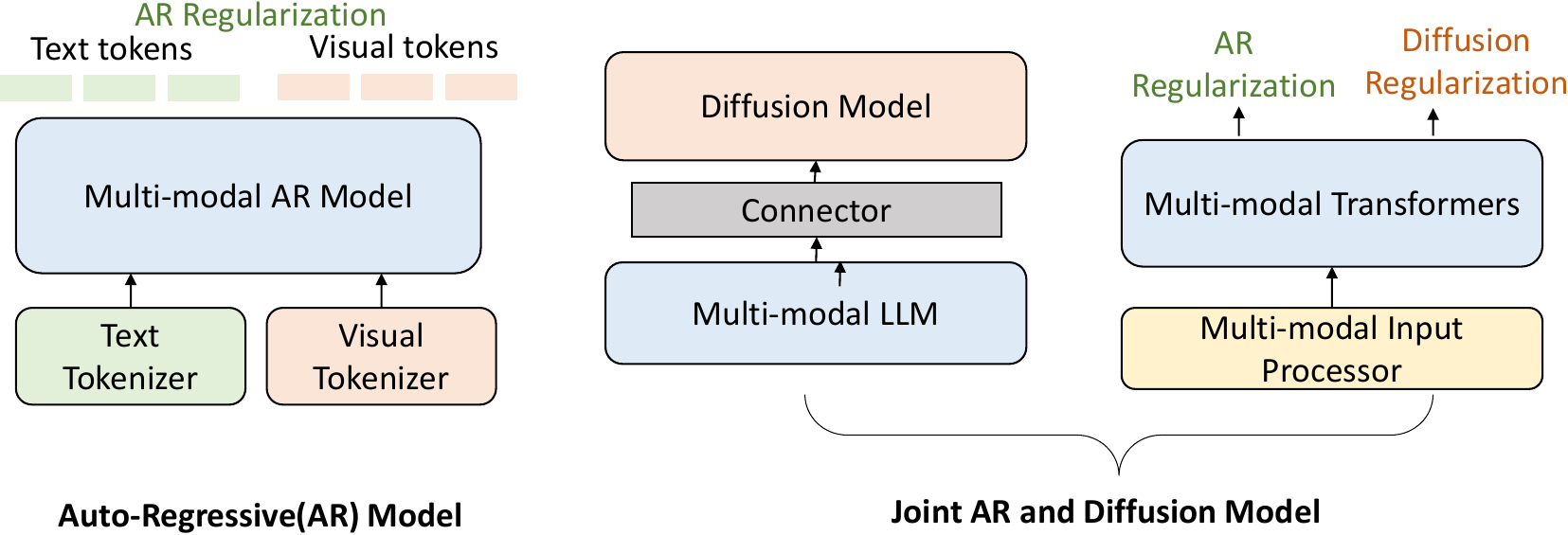}
    \caption{Possible unified multi-modal understanding and generation frameworks with different probabilistic modeling methods. }
    \label{fig:ar_diffusion}
\end{figure}

The success of multi-modal large-language models has clearly shown the great power of autoregressive modeling for multi-modal understanding and text generation, so we believe the autoregressive method should be included. Then, the next question is how we enable the model with visual generation ability. Based on existing works in Sec.~\ref{sec:MLLM} and Sec.~\ref{sec:diffusion}, we provide the possible methods in Fig.~\ref{fig:ar_diffusion}, where we present the autoregressive model and the joint autoregressive and diffusion model. Next, we will elaborate on them in detail.

\subsubsection{Autoregressive (AR) Model} 
Although diffusion models have become dominant in visual generation, there are still some recent attempts~\cite{zhu2023vlgpt,chameleon,llamagen,zhan2024anygpt,sun2024emu2,wang2024emu3, wu2025janus} on generating visual content in an autoregressive manner. These works will first try to map the input images and text into discrete tokens, respectively. Particularly, the images are discretized with visual tokenizers such as VQGAN or VQ-VAE. Then the mixed text and visual tokens will be fed into a multi-modal autoregressive model. After that, the model will output the mixed text and visual tokens. Also, some special tokens such as $<soi>, <eoi>$ are used to indicate the start of the image tokens and the end of the image tokens. Then the generated text tokens will deliver how the model understands the input multi-modal information, and the visual tokens will be sent to the decoder of the VQ-VAE or VQGAN to reconstruct images. Therefore, the autoregressive model can be used for both understanding and visual generation.

\noindent \textbf{Remark.} Despite these efforts, the autoregressive method is far from perfect --- it basically assumes the existence of a causal structure and causal attention, where previous tokens are used to predict next tokens. However, this is not suitable for image generation because it is difficult to determine, which visual token should be the first and which one should be the last. Therefore, a recent work VAR~\cite{nextscale} tries to use the next-scale prediction paradigm to generate images, where the lower-resolution images are regarded as previous tokens to predict (next) higher-resolution images. Unfortunately, the scaling ability is still not verified in multi-modal understanding and generation, and the model achieves a 1.73 FID score on the ImageNet~\cite{deng2009imagenet} benchmark for generation, falling behind the diffusion model~\cite{yao2025reconstruction} which has a 1.35 FID score. \zyw{In general, joint AR and diffusion models outperform unified AR models on visual generation tasks. For instance, EMU3~\cite{wang2024emu3} and Janus-Pro~\cite{chen2025januspro}, both unified AR models, achieve 0.66 and 0.80 on the GenEval benchmark, respectively. In contrast, joint AR-diffusion models such as Mogao~\cite{MOGAO} and Bagel~\cite{BAGEL} reach 0.89 and 0.88, demonstrating the advantages of combining AR and diffusion components for visual generation.}

\subsubsection{Joint Autoregressive and Diffusion Model}
Considering the impressive visual generation ability of the diffusion model, a more natural way for unified multi-modal understanding and generation is to combine the autoregressive and diffusion models. In Fig.~\ref{fig:ar_diffusion}, we present two kinds of possible frameworks. 

The first one is that we have a pretrained diffusion model for visual generation and a multi-modal LLM for multi-modal understanding. \hb{We then connect these two components, forming what we call Connector-based Joint Models.} Regarding how to connect these two parts, many existing works~\cite{visualgpt,hugginggpt,tool-LLM} directly use the LLM as the controller and the diffusion model as a tool for visual generation, which is a common paradigm in tool learning. Although works like tool learning can enable the models with visual generation abilities, they easily suffer from generation failure when meeting multi-modal generation conditions. For example, when we want to generate ``a specific girl (described with a given image) and a specific dog (described with a given image) playing on the grass'', the tools available are only SOTA text-to-image models. They will fail to guarantee that the specific girl and dog occur in the generated image. In fact, there are many conditions that cannot be described with only text, and this kind of tool-learning method will fail. To tackle the problem, a more advanced way is to train a learnable connector~\cite{kosmos-G,codi2,ge2024seed, chen2025blip3o}, which aligns the diffusion model and the multi-modal LLM in the same space, similar to the training paradigm of the alignment module in multi-modal LLM. The alignment process enables the diffusion model to receive the LLM output multi-modal embeddings as conditions instead of pure text descriptions, thus achieving multi-modal generation. However, this paradigm inherits the limitations of alignment architecture. The multi-modal LLM and the diffusion model are pretrained respectively. The performance of the unified model will be limited by each model. Additionally, from an intuitive perspective, multi-modal understanding and multi-modal generation should not be independent tasks but rather two related tasks that could share knowledge. \hb{To train such a model, both the MLLM and the diffusion model can be frozen, and only the connector is trained. This maximally preserves the capabilities of the two models, but the information bottleneck between them can be particularly severe. Alternatively, one or both of the models can be included in training, but this requires a larger amount of data and computational resources to ensure that the original abilities of the models are not compromised. For example, in Qwen-Image~\cite{wu2025qwen}, the MLLM is kept frozen while the diffusion model is trained on a large dataset. This preserves the full capability of the MLLM while endowing it with strong generative ability.}

The second possible model is a unified multi-modal-transformer framework as shown in Fig.~\ref{fig:ar_diffusion}, where we do not rely on two pretrained models, but try to use a single model trained with both diffusion and autoregressive regularizations, \hb{which we refer to as Autoregressive-Diffusion Joint Models.} The multi-modal input processor will first transform the multi-modal data into sequences that can be received by the transformers. Then the multi-modal transformer will try to learn the multi-modal knowledge for both understanding and generation. \hb{Specifically, the training objectives are designed differently for each modality: text prediction uses an autoregressive regularization (computed token-wise), while image prediction uses a diffusion regularization (computed over the entire image, covering multiple patches). During inference, the model dynamically switches between language modeling and diffusion modes. In language modeling mode, it samples tokens sequentially; upon generating the BOI token, it switches to diffusion mode, appending a sequence of pure noise patches corresponding to the target image size, and gradually generates the image through T-step denoising iterations. At each step, the model predicts the noise based on the current image representation and updates the patch sequence until denoising is complete. The EOI token is then appended, and the model switches back to language modeling mode.}
Note that this is a transformer-like model but not necessarily an LLM. This is because when using transformers to generate visual content, the full-attention mechanism is usually adopted. In contrast, the attention mechanism adopted by LLM is causal and uni-directional. Therefore, an adaptive or mixed attention mechanism might be designed. This perspective is verified in TransFusion~\cite{transfusion} and Show-o~\cite{showo}. The difference between Transfusion and Show-o mainly lies in the diffusion model, where TransFusion adopts continuous diffusion that is similar to current visual diffusion models, but Show-o adopts masked generative modeling~\cite{magvit}, which could be regarded as discrete diffusion regularization. Therefore, Show-o still relies on a pixel-level visual tokenizer for image generation but might trade off some understanding ability. Additionally, these two works are primary attempts at combining autoregressive and diffusion modeling methods in a single transformer-like model. There still exist several open problems regarding what the model architecture should be like, such as the multi-modal input processor or the transformer-like model, which we will discuss next.

\subsection{Model Architecture}
Compared to previous multi-modal LLM or Diffusion models that only focus on one task, i.e., generation or understanding, the unified model itself should support multiple objectives. When it comes to understanding, the model should have the ability of conceptual abstraction and associative reasoning. In contrast, when it comes to visual generation, besides the overall concepts and their relations, pixel-level details are also important. Therefore, the unified model architecture design might be different from that of previous single-objective models. Next, we mainly discuss the possible architectures of the multi-modal input processor and the multi-modal transformers.

\subsubsection{Multi-modal input processor} 
\begin{figure}
    \centering
    \includegraphics[width=0.65\linewidth]{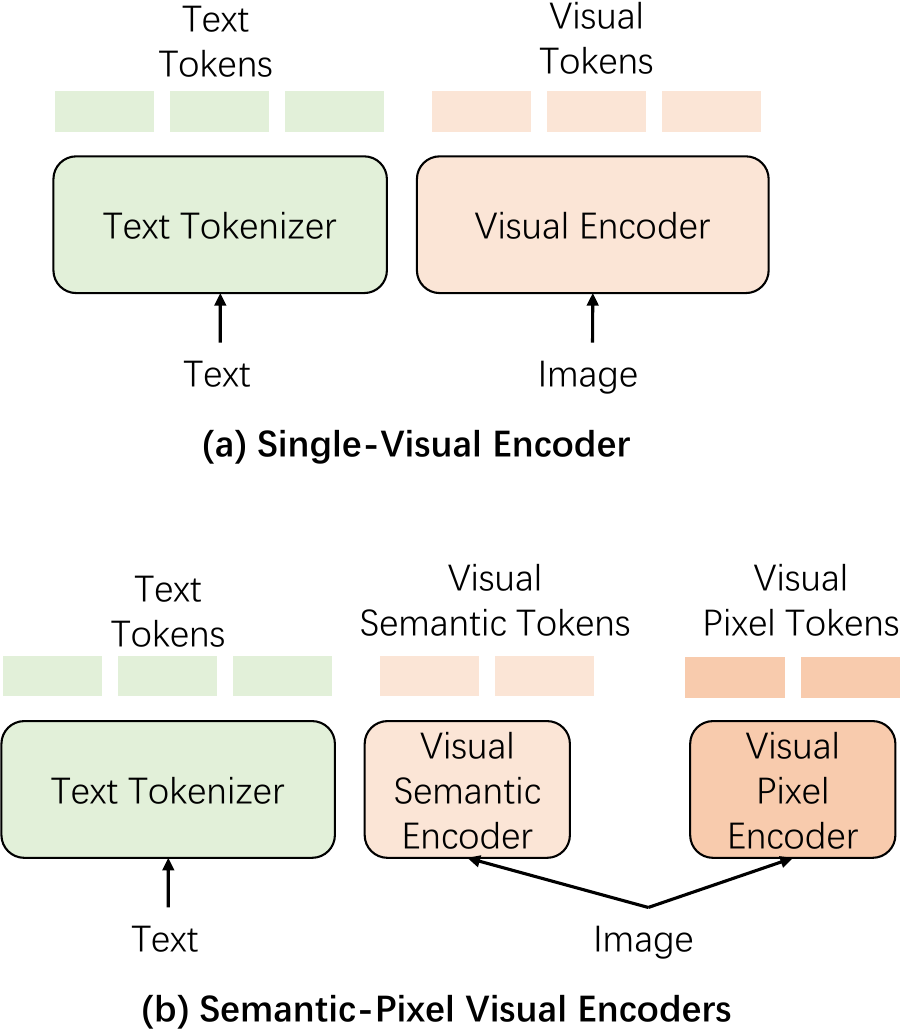}
    \caption{Possible frameworks of the multi-modal input processor for unified multi-modal understanding and generation models.}
    \label{fig:input_processor}
\end{figure}

To tackle the multi-modal input text and images, two possible input processors are presented in Fig.~\ref{fig:input_processor}. Text is consistently tackled by a text tokenizer. However, there are some differences in the visual input. In Fig.~\ref{fig:input_processor}(a), we show the visual processor adopted by most early works, where a single visual encoder is used to process the images. Considering that the visual tokens should support the pixel-level visual generation task, early works~\cite{chameleon,showo,transfusion} generally adopt the single pixel-level (or patch-level) visual tokens (e.g., VQVAE). The pixel-level tokens bring challenges to the multi-modal transformer, requiring it not only to capture the relations between image patches for visual generation but also to visual abstract reasoning ability for understanding. In contrast, a possible alternative multi-modal input processor is presented in Fig.~\ref{fig:input_processor}(b). For each image, we respectively use a semantic encoder (e.g., CLIP-ViT) and a pixel-level encoder (e.g., VQVAE) to obtain both semantic and pixel tokens. Janus~\cite{wu2025janus} was the first to adopt this architecture. It introduced two separate visual encoding paths: a semantic encoder for extracting visual features in understanding tasks, and a pixel-level encoder for encoding images in generation tasks. Subsequent works, such as UniToken~\cite{jiao2025unitoken}, further explored directly concatenating features of the two encoders along the sequence dimension, allowing the model to receive both types of features simultaneously for understanding and generation tasks. By using a dual-encoder approach, models can leverage both low-level pixel information and high-level semantic information, which better enhances performance on both understanding and generation tasks. Consequently, most recent works adopt this architecture. Moreover, it is a more flexible way to conduct some adaptive token selection from the semantic and pixel tokens for fine-grained understanding. We believe this would result in interesting research work.

\subsubsection{Multi-modal Transformer} 
\begin{figure}
    \centering
    \includegraphics[width=\linewidth]{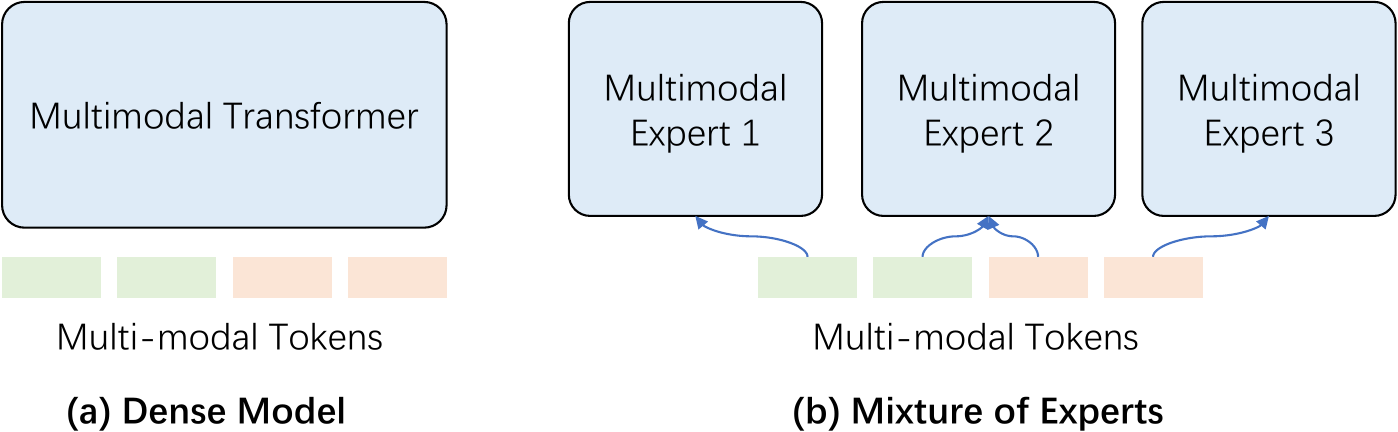}
    \caption{Possible architectures of the multi-modal transformer.}
    \label{fig:transformer}
\end{figure}

After discussing how to tackle the multi-modal input information, the next key component is the multi-modal transformer, which captures the complex relations among and within modalities. As shown in Fig.~\ref{fig:transformer}, on the left is a dense model, where one unified transformer is used for both multi-modal understanding and generation~\cite{lin2023video,chen2024internvl}. Considering that understanding and generation might share some knowledge but their objectives are not exactly the same, it is a natural idea to utilize the mixture of experts~\cite{jacobs1991moe} in multi-task learning as shown in (b). On the right of the figure, some of the experts share the knowledge of understanding and generation, e.g., concepts and their relations, some of the experts are good at analyzing visual details for visual generation, and other experts are good at conducting reasoning for better understanding. LlamaFusion~\cite{shi2024llamafusion} and BAGEL~\cite{BAGEL} have made preliminary explorations in this area, both using only two experts and employing hard routing. In LlamaFusion, which uses a single visual encoder, one expert is responsible for processing text tokens, while the other handles visual tokens. In contrast, BAGEL, which adopts semantic-pixel visual encoders, assigns one expert to process text tokens and visual semantic tokens, and the other to handle visual pixel tokens. Both works find that their architectures outperform dense models, indicating that unified models still face optimization challenges arising from task-specific or modality-specific learning objectives.

\hb{In Table~\ref{tab:vlm_comparison}, we present the performance of several recent unified models. Due to large differences in model size and training data volume, a fair comparison is difficult. Regarding architecture choice: currently, there are still no large-scale Autoregressive Models trained with massive data. The latest Skywork UniPic demonstrates strong capabilities in generation and editing, but its performance on understanding tasks is not reported. In the Connector-based Joint Models category, MetaQueries, BLIP3o, and Qwen-Image all adopt Qwen2.5-vl-7B as the MLLM, resulting in similar performance on understanding tasks. However, the success of Qwen-Image indicates that increasing the scale of the diffusion model and enlarging the training dataset can significantly boost performance in generation and editing tasks. In the Autoregressive-Diffusion Joint Models category, BAGEL leverages the largest model and dataset, making it a strong competitor to Qwen-Image. Regarding the choice of visual encoder: most recent models adopt the dual encoder (Semantic-Pixel Visual Encoders) architecture, which benefits both understanding and generation tasks. Since models using MoE are still limited, it remains unclear whether MoE brings significant advantages. We hope that future work will explore this direction further.}

In this section, we provide a discussion of the unified model of multi-modal generation and multi-modal understanding, from both the probabilistic modeling methods and the model architectures. Though the discussed techniques can combine with each other to form more architectures as well, there are very few attempts at the unified model design, making us believe the inspirations of many future works brought by the discussions above.

\begin{table*}[!t]
\caption{Overview of multi-modal LLM, diffusion, and unified models in this paper.}
\label{tab:overview}
\centering
\begin{tabular}{llllllc}
\toprule
\textbf{Model}   & \textbf{Institution}   & \textbf{Type}  & \textbf{Classification}    & \textbf{Publication} & \textbf{Year} & \textbf{Parameters} \\
\midrule
\multicolumn{7}{c}{\textbf{Multi-modal LLM (MLLM)}}                                         \\ \hline
LLaVA~\cite{llava} & Microsoft
& Image LLM   & Alignment          & NeurIPS              & 2024          & 13B \\
BLIP-2~\cite{blip2} & Salesforce
& Image LLM   & Alignment          & ICML                 & 2023          & 12B \\
MiniGPT-4~\cite{zhu2023minigpt}    & KAUST
& Image LLM   & Alignment          & ICLR                 & 2024          & 7B \\
Qwen-VL~\cite{bai2023qwen} & Alibaba
& Image LLM   & Alignment          & ArXiv                & 2023          & 7B \\
Flamingo~\cite{alayrac2022flamingo} & DeepMind
& Image LLM   & Alignment          & NeurIPS              & 2025          & 3B \\
Fuyu~\cite{adept_fuyu_8b} & Adept
& Image LLM   & Early-Fusion       & -                    & 2023          & 8B \\
Gemini~\cite{team2023gemini} & Google
& Image LLM   & Early-Fusion       & ArXiv                & 2023          & - \\
Claude3~\cite{claude3} & Anthropic 
& Image LLM     & Early-Fusion      & -                   & 2024          & - \\
VideoChat~\cite{2023videochat} & Shanghai AI Lab
& Video LLM   & Alignment          & ArXiv                & 2023          & 7B \\
VideoLLaMA~\cite{zhang2023video} & Alibaba
& Video LLM   & Alignment          & EMNLP                & 2023          & 7B \\
VideoLLaMA2~\cite{damonlpsg2024videollama2} & Alibaba
& Video LLM   & Alignment          & ArXiv                & 2024          & 7B \\
Video-ChatGPT~\cite{Maaz2023VideoChatGPT} & MBZUAI
& Video LLM   & Alignment          & ACL                  & 2023          & 7B \\
LLaVA-OneVision~\cite{LLaVA-OneVision}  & ByteDance
& Video LLM   & Alignment          & TMLR                 & 2024          & 7B \\
MiniCPM-V~\cite{yao2024minicpm}    & OpenBMB    
& Video LLM   & Alignment          & ArXiv                & 2024          & 8B \\
VILA-1.5~\cite{lin2023vila}    & NVIDIA    
& Video LLM   & Alignment          & ArXiv                & 2023          & 7B \\ 
Pengi~\cite{deshmukh2023pengi} & Microsoft
& Speech LLM  & Alignment & NeurIPS    & 2023     & 1B \\
Salmonn~\cite{tang2023salmonn} & ByteDance
& Speech LLM  & Alignment  & ICLR      & 2024        & 13B \\
Qwen-Audio~\cite{chu2023qwen} & Alibaba
& Speech LLM  & Alignment  & ArXiv    & 2023        & 7B \\
OSUM~\cite{geng2025osum} & NPU
& Speech LLM  & Alignment  & ArXiv    & 2025        & 7B \\
VALL-E~\cite{chen2024vall} & Microsoft
& Speech LLM  & Early-Fusion  & ArXiv           & 2025        & 7B \\
SpeechGPT~\cite{zhang2023speechgpt} & Fudan University
& Speech LLM  & Early-Fusion  & EMNLP           & 2023        & 7B \\
AudioPaLM~\cite{rubenstein2023audiopalm} & Google
& Speech LLM  & Early-Fusion  & ArXiv           & 2023        & 8B \\
\hline
\multicolumn{7}{c}{\textbf{Diffusion Model}}                                                        \\ \hline
GLIDE~\cite{nichol2021glide}      & OpenAI
& Text-to-Image  & Pixel-Based        & ICML                 & 2022          & 5B \\
Imagen~\cite{saharia2022photorealistic}      & Google    
& Text-to-Image  & Pixel-Based        & NeurIPS              & 2022          & 3B \\
Stable Diffusion~\cite{stablediffusion} & LMU
& Text-to-Image  & Latent-Based       & CVPR                 & 2022          & 1B \\
DALL-E2~\cite{ramesh2022hierarchical} & OpenAI
& Text-to-Image  & Latent-Based       & ArXiv                & 2022          & 6B \\
DiT~\cite{peebles2023scalable}      & Meta         
& Text-to-Image  & Latent-Based       & ICCV                 & 2023          & 1B \\
PixArt-$\alpha$~\cite{chen2023pixart}  & Huawei      
& Text-to-Image  & Latent-Based       & ICLR                 & 2025          & 1B \\
Text2Video-Zero~\cite{khachatryan2023t2vzero}  & Picsart AI
& Text-to-Video  & Latent-Based    & ICCV                 & 2023          & 1B \\
Latent-Shift~\cite{an2023latentshift}    & Meta
& Text-to-Video  & Latent-Based    & ArXiv                & 2023          & 2B \\
VDM~\cite{ho2022videodiffusionmodels}        & Google
& Text-to-Video  & Latent-Based      & NeurIPS              & 2022         & - \\
Make-a-video~\cite{singer2024makeavideo}    & Meta
& Text-to-Video  & Latent-Based      & ICLR                 & 2024          & 10B \\
AnimateDiff~\cite{guo2024animatediff}      & Shanghai AI Lab
& Text-to-Video   & Latent-Based      & ICLR                 & 2024          & 1B \\
Latte~\cite{ma2024latte}     & Shanghai AI Lab      
& Text-to-Video  & Latent-Based      & TMLR                 & 2025          & 1B \\
CogVideo~\cite{hong2023cogvideo}   & Z.AI         
& Text-to-Video  & Latent-Based      & ICLR                 & 2023          & 15B \\
Wan~\cite{wan2025wan}   & Alibaba       
& Text-to-Video  & Latent-Based      & ArXiv                 & 2025          & 14B\\
HunyuanVideo~\cite{kong2024hunyuanvideo}   & Tencent         
& Text-to-Video  & Latent-Based      & ArVix                 & 2024          & 13B \\
Vidu~\cite{bao2024vidu}        & Shengshu    
& Text-to-Video  & Latent-Based      & ArXiv                 & 2024          & - \\ 
DiffWave~\cite{kong2020diffwave}  & Baidu
& Text-to-Speech  & Vocoder  & ICLR  & 2021  & 6M \\
WaveGrad~\cite{chen2020wavegrad,chen2021wavegrad}  & Google
& Text-to-Speech  & Vocoder  & ICLR  & 2021  & 23M \\
Grad-TTS~\cite{popov2021grad}  & Huawei
& Text-to-Speech  & Acoustic Model  & ICML  & 2021  & 30M \\
Diff-TTS~\cite{jeong2021diff}  & Neosapience
& Text-to-Speech  & Acoustic Model  & Interspeech  & 2021  & 13M \\
\hline
\multicolumn{7}{c}{\textbf{Unified Model}}  \\ \hline
VL-GPT~\cite{zhu2023vlgpt}     & Tencent         
& Unified Model    & Autoregressive     & ArXiv                & 2023         & 8B  \\
Chameleon~\cite{chameleon}       & Meta       
& Unified Model   & Autoregressive     & ArXiv                & 2024          & 7B  \\
Emu2~\cite{sun2024emu2}       & BAAI       
& Unified Model    & Autoregressive     & CVPR                 & 2024          & 37B  \\
Emu3~\cite{wang2024emu3}     & BAAI        
& Unified Model    & Autoregressive     & ArXiv                & 2024          & 8B  \\
LlamaGen~\cite{llamagen}      & ByteDance        
& Unified Model    & Autoregressive     & ArXiv                & 2024          & 3B  \\
AnyGPT~\cite{zhan2024anygpt}   & Shanghai AI Lab             
& Unified Model    & Autoregressive     & ACL                  & 2024          & 8B  \\
Janus~\cite{wu2025janus}      & DeepSeek       
& Unified Model    & Autoregressive     & CVPR                  & 2025          & 1B  \\
Janus-Pro~\cite{chen2025januspro}    & DeepSeek         
& Unified Model    & Autoregressive     & ArXiv                  & 2025         & 7B  \\
Skywork UniPic~\cite{wang2025skywork}    & Skywork         
& Unified Model    & Autoregressive     & ArXiv                  & 2025         & 2B  \\
VisualGPT~\cite{visualgpt}    & Microsoft      
& Unified Model    & Joint AR-Diffusion & ArXiv                & 2023         & -  \\
HuggingGPT~\cite{hugginggpt}   & Microsoft          
& Unified Model    & Joint AR-Diffusion & NeurIPS              & 2024          & -  \\
MLLM-Tool~\cite{tool-LLM}    & Meituan         
& Unified Model    & Joint AR-Diffusion & WACV                 & 2025          & 13B  \\
Kosmos-G~\cite{kosmos-G}      & Microsoft       
& Unified Model    & Joint AR-Diffusion & ICLR                 & 2024          & 2B  \\
CoDi-2~\cite{codi2}        & Microsoft     
& Unified Model    & Joint AR-Diffusion & CVPR                 & 2024          & 8B  \\
Seed-X~\cite{ge2024seed}   & Tencent          
& Unified Model    & Joint AR-Diffusion & ArXiv                & 2024          & 13B  \\
MetaQuery~\cite{MetaQuery}  & Meta     
& Unified Model    & Joint AR-Diffusion & ArXiv                & 2025          & 7B  \\
BLIP3o~\cite{chen2025blip3o}  & Salesforce     
& Unified Model    & Joint AR-Diffusion & ArXiv                & 2025          & 8B  \\
OmniGen2~\cite{wu2025omnigen2}  & BAAI     
& Unified Model    & Joint AR-Diffusion   & ArXiv                & 2025          & 7B  \\
Qwen-Omni~\cite{xu2025qwen2,xu2025qwen3}  & Alibaba     
& Unified Model    & Joint AR-Diffusion   & ArXiv                & 2025          & 30B  \\
Ming-Omni~\cite{ai2025ming}  & Ant Group     
& Unified Model    & Joint AR-Diffusion   & ArXiv                & 2025          & 7B  \\
TransFusion~\cite{transfusion}    & Meta     
& Unified Model    & Joint AR-Diffusion & ICLR                 & 2025          & 7B  \\
Show-o~\cite{showo}           & NUS     
& Unified Model    & Joint AR-Diffusion & ICLR                 & 2025          & 1B  \\
Show-o2~\cite{xie2025showo2}           & NUS     
& Unified Model    & Joint AR-Diffusion            & ArXiv                 & 2025          & 7B  \\
LlamaFusion~\cite{shi2024llamafusion}     & Meta            
& Unified Model    & Joint AR-Diffusion & Arxiv                & 2024         & 8B  \\
Mogao~\cite{MOGAO}        & ByteDance        
& Unified Model    & Joint AR-Diffusion & Arxiv                & 2025         & 7B  \\
BAGEL~\cite{BAGEL}        & ByteDance        
& Unified Model    & Joint AR-Diffusion & Arxiv                & 2025         & 7B  \\
\bottomrule
\end{tabular}
\end{table*}

\begin{table*}[t]

\caption{Comparison of recent multi-modal models across understanding, generation, and editing benchmarks.}
\label{tab:vlm_comparison}
\centering
\scriptsize
\setlength{\tabcolsep}{3pt}
\renewcommand{\arraystretch}{1.15}
\begin{tabular}{lccccccccccccc}
\toprule
\multirow{2}{*}{Model} & \multirow{2}{*}{Date} & \multirow{2}{*}{Params} & \multirow{2}{*}{Data} & \multirow{2}{*}{Dual Encoder} & \multirow{2}{*}{MoE} & \multicolumn{4}{c}{Understanding} & \multicolumn{2}{c}{Generation} & \multicolumn{2}{c}{Editing} \\
\cmidrule(lr){7-9} \cmidrule(lr){10-12} \cmidrule(lr){13-14}
 &  &  &  &  &  & MMBench & MMMU & MM-Vet & WISE & GenEval & DPGBench & ImgEdit & GEdit-Bench-EN \\
\midrule
\rowcolor{gray!10}
GPT-4o & 2025.3 & - & - & - & - & 86.0 & 70.7 & - & 0.80 & 0.89 & 86.23 & 4.20 & 7.53 \\
\midrule
\multicolumn{14}{l}{\textbf{Autoregressive Models}} \\
\midrule
Emu3~\cite{wang2024emu3} & 2024.9 & 8B & - & $\times$ & $\times$ & 58.5 & 31.6 & 37.2 & 0.39 & 0.66 & 80.6 & - & - \\
Janus-Pro~\cite{chen2025januspro} & 2025.1 & 7B & 144M & \checkmark & $\times$ & 79.2 & 41.0 & 50.0 & 0.35 & 0.80 & 84.19 & - & - \\
Skywork UniPic~\cite{wang2025skywork} & 2025.8 & 2B & 130M & \checkmark & $\times$ & - & - & - & - & 0.86 & 85.50 & 3.49 & 5.83 \\
\midrule
\multicolumn{14}{l}{\textbf{Connector-based Joint Models}} \\
\midrule
MetaQueries~\cite{MetaQuery} & 2025.4 & 7B+1.6B & 25M & \checkmark & $\times$ & 83.5 & 58.6 & 66.6 & 0.55 & 0.80 & 82.05 & - & - \\
BLIP3o~\cite{chen2025blip3o} & 2025.5 & 7B+1.4B & 25M & \checkmark & $\times$ & 83.5 & 50.6 & 66.6 & 0.62 & 0.84 & 81.6 & - & - \\
OmniGen2~\cite{wu2025omnigen2} & 2025.6 & 3B+4B & 66M & \checkmark & $\times$ & 79.1 & 53.1 & 61.8 & - & 0.80 & 83.57 & 3.44 & 6.42 \\
Qwen-Image~\cite{wu2025qwen} & 2025.8 & 7B+20B & $>$1000M & \checkmark & $\times$ & 83.5 & 58.6 & 67.1 & - & 0.87 & 88.32 & 4.27 & 7.56 \\
\midrule
\multicolumn{14}{l}{\textbf{Autoregressive-Diffusion Joint Models}} \\
\midrule
Mogao~\cite{MOGAO} & 2025.5 & 7B & - & \checkmark & \checkmark & 75.0 & 44.2 & - & - & 0.89 & 84.33 & - & - \\
BAGEL~\cite{BAGEL} & 2025.5 & 14B & 1600M & \checkmark & \checkmark & 85.0 & 55.3 & 67.2 & 0.52 & 0.88 & 85.07 & 3.20 & 6.52 \\
Show-o2~\cite{xie2025showo2} & 2025.6 & 7B & 66M & \checkmark & $\times$ & 79.3 & 48.9 & - & - & 0.76 & 86.14 & - & - \\
\bottomrule
\end{tabular}
\end{table*}

\section{Datasets}
\label{sec:data}
After discussing the multi-modal understanding and generation models, multi-modal text-image and text-video datasets are also important to implement multi-modal generative AI\cite{zhu2015multimedia}. In this section, we will review the literature on the datasets for training multi-modal generative AI models.
Based on the differences in data types, we divide the datasets into three categories: caption, conversation, and reasoning.
In addition, many multi-modal large foundation models choose to collect the aforementioned types of data for integration and construct their own datasets. Therefore, we denote these datasets as the integration datasets.

\begin{table*}[ht]
\small
    \centering
    \caption{Common datasets}
    \begin{tabular}{ccccc}
    \toprule
        Dataset type  & Modalities& Datasets  \\
        \midrule
        \multirow{2}{*}{Captions}
        &Text-Image&
        SBU Captions~\cite{ordonez2011im2text}, MSCOCO~\cite{lin2014microsoft}, CC-3M~\cite{sharma2018conceptual}, LAION~\cite{schuhmann2022laion}, MINT-1T~\cite{awadalla2024mint}\\
        &Text-Video&WebVid~\cite{bain2021frozen}, InternVid~\cite{wanginternvid}, HD-VG-130M~\cite{wang2023videofactory}, YouCook2~\cite{zhou2018towards}, TextVR~\cite{wu2025large}\\
        \midrule
        \multirow{2}{*}{Conversation}
        &Text-Image&VQAv2~\cite{goyal2017making}, GQA~\cite{hudson2019gqa}, OK-VQA~\cite{marino2019ok}, AOK-VQA~\cite{schwenk2022okvqa}, OCR-VQA~\cite{mishra2019ocr}, TextVQA~\cite{singh2019towards}\\
        &Text-Video&
        TGIF-QA~\cite{jang2017tgif}, WebVidQA~\cite{yang2021just}, EgoQA~\cite{grauman2022ego4d}\\
        \midrule
         \multirow{2}{*}{Reasoning}
         &Text-Image&CLEVR~\cite{johnson2017clevr}, VisualMRC~\cite{tanaka2021visualmrc}\\
         &Text-Video&NExT-QA~\cite{xiao2021next}, CLEVRER~\cite{yi2020clevrer}\\
         \midrule
         \multirow{2}{*}{Intergration}
         &Text-Image&LLaVA-Instruct~\cite{llava}\\
         &Text-Video\&Image&Video-LLaVA~\cite{lin2023video}, VideoChat2~\cite{li2024mvbench}, VideoLLaMa2~\cite{damonlpsg2024videollama2}\\
    \bottomrule
    \end{tabular}
    \label{tab:generator}
\end{table*}

\subsection{Caption Datasets}
The caption dataset aims to improve basic visual and temporal description capabilities for multi-modal LLMs and provide the mapping relationship for text-to-image and text-to-video models. Commonly used text-to-image datasets include SBU Captions~\cite{ordonez2011im2text},  MSCOCO~\cite{lin2014microsoft}, Conceptual Captions (CC-3M)~\cite{sharma2018conceptual}, and LAION~\cite{schuhmann2022laion}. The size of these datasets ranges from 328K to 5B. Recently, MINT-1T has been proposed, comprising one trillion text tokens and three billion images~\cite{awadalla2024mint}, a 10x scale-up from existing open-source datasets, and it includes previously untapped sources such as PDFs and ArXiv papers. Text-to-video datasets include WebVid~\cite{bain2021frozen}, InternVid~\cite{wanginternvid}, HD-VG-130M~\cite{wang2023videofactory}, YouCook2~\cite{zhou2018towards}, and TextVR~\cite{wu2025large}.

The caption datasets mainly serve in the following two aspects, i.e., (i) provide knowledge for the training of generation models to generate images or videos based on the input text embedding, and (ii) use text-image datasets to align the image modality with the multi-modal LLM for understanding inputs.

\subsection{Conversation Datasets}
The conversation dataset aims at enhancing multi-modal LLMs' capabilities for single-turn and multi-turn conversations when asking questions about the input image or video. Normally, a diverse set of questions would be asked about the visual content of the image and the video, including the object types, counting the objects, object actions, object locations, event moment, event duration, and relative positions between objects. With simple formatting reorganization, many visual QA datasets could be directly constructed as conversation datasets for multi-modal LLM training. These include basic VQA (VQAv2~\cite{goyal2017making}, GQA~\cite{hudson2019gqa}), knowledge-based VQA (OK-VQA~\cite{marino2019ok}, AOK-VQA~\cite{schwenk2022okvqa}), OCR-based VQA (OCR-VQA~\cite{mishra2019ocr}, TextVQA~\cite{singh2019towards}) and VideoQA (TGIF-QA~\cite{jang2017tgif}, WebVidQA~\cite{yang2021just}, and egocentric VQA from Ego4D~\cite{grauman2022ego4d}), which can not only improve the visual QA capabilities for multi-modal LLMs in conversations but also help the models to learn more visual and temporal knowledge.

\subsection{Reasoning Datasets}
The above two types of datasets mainly focus on the visual content itself, normally lacking in-depth reasoning questions. Meanwhile, the reasoning datasets focus on enhancing multi-modal LLMs for diverse reasoning capacities, which normally require a step-by-step reasoning process by following rigorous logic. These include spatial reasoning (CLEVR~\cite{johnson2017clevr}), reading comprehension (VisualMRC~\cite{tanaka2021visualmrc}), temporal reasoning (NExT-QA~\cite{xiao2021next}), and spatiotemporal reasoning (CLEVRER~\cite{yi2020clevrer}).

\subsection{Integration Datasets}
Due to the strong generalization ability of multi-modal LLMs, their training data is not limited to only one single task, such as caption, conversation, or reasoning, instead requiring comprehensive pretraining for both simple and complex visual modal tasks. Therefore, many multi-modal large model works often do not use a single visual task dataset. Instead, they select subsets of several datasets from each category mentioned above for integration and adjustment, forming instruction training data that employs both image and video data for different visual modal tasks.
For visual instruction tuning, LLaVA~\cite{llava} is the first multi-modal LLM, which i) leverages text-only GPT-4~\cite{GPT4V} to expand the existing bounding box, and ii) employs caption dataset (e.g., MSCOCO~\cite{lin2014microsoft}) as multi-modal instruction tuning data.
In addition, Liu et al. propose LLaVA-Instruct, which is built on a subset of the CC-3M dataset and contains 58k in conversations, 23k in detailed descriptions, as well as 77k in complex reasoning records.
Following the development of visual instruction tuning, many video LLMs such as Video-LLaVA~\cite{lin2023video}, VideoChat2~\cite{li2024mvbench}, and VideoLLaMa2~\cite{damonlpsg2024videollama2}, are proposed, utilizing the combination of caption, conversation, and reasoning datasets under both text-image and text-video modalities.

\section{Future Directions}
\label{sec:future}

Last but not least, we explore challenging problems deserving further investigation and share our insights on promising future directions for multi-modal generative AI.

\subsection{Unified Model for Video Understanding and Generation}

In Section IV, we primarily discuss the unified models for image understanding and generation. Given the large amount of video data in the wild, we believe there will be an urgent need to extend the unification to videos~\cite{zhu2024multi,you2024towards,jin2024mtartgpt}. Among the three architectures introduced in Fig.~\ref{fig:ar_diffusion}, bridging the multi-modal LLM and video diffusion model with a connector~\cite{Next-gpt, X-VILA} can be achieved in a way similar to images. 
However, adapting the other two architectures to videos faces significant challenges due to i) the increased computational demands caused by longer sequences, as well as ii) the difficulty in learning spatiotemporal cues.
For instance, in an autoregressive model, encoding individual video frames separately using a 2D visual tokenizer fails to capture the essential temporal motion information. VideoPoet~\cite{kondratyuk2024videopoet}, which employs a 3D video tokenizer~\cite{magvit-v2}, encodes a 17-frame video (spanning 2.125 seconds) into 1280 tokens, limiting its ability to generate longer videos. VideoLaViT~\cite{Video-lavit} introduces an efficient video representation model by decomposing videos into keyframes and temporal motions, training separate tokenizers for each of them, which significantly improves computational efficiency. However, the training cost is still too high when scaling to the large amount of web-scale video data.
Similarly, using a single model trained with both diffusion and autoregressive regularizations also encounters the same challenges, where modeling complex relations such as causal attention and spatiotemporal attention within the model remains unexplored. Therefore, it deserves more effort in advancing unified generative AI for video understanding and generation.

\subsection{Benchmark for the Unification}

On the one hand, despite some pioneering work on studying unified models~\cite{transfusion,showo} for understanding and generation, the corresponding evaluations are conducted separately in a non-unified way. For instance, existing works use specific benchmarks for understanding tasks, such as Flickr30k~\cite{young2014image} and VQAv2~\cite{goyal2017making}, while relying on different benchmarks for generation tasks, such as MSCOCO~\cite{lin2014microsoft} and GenEval~\cite{ghosh2024geneval}. On the other hand, a unification benchmark offers the advantage of unified metrics and rankings, providing a more comprehensive and fair assessment of model performance across both tasks. However, designing such a benchmark is challenging, as it requires a vast amount of visual data with human annotations in various forms, including labels, rankings, and natural language descriptions. More importantly, the evaluation should ideally reflect the mutual promotion between understanding and generation. In summary, the challenges for creating a unification benchmark are threefold, 
\begin{enumerate}
    \item \textbf{Dataset construction.} The visual data should be representative, diverse, and abundant, with high-quality annotations for multiple tasks.
    \item \textbf{Ranking criteria.} Models should be ranked based on a combination of understanding and generation metrics, ensuring a balanced evaluation of both capabilities.
    \item \textbf{Mutual promotion.} The benchmark should include datasets or tasks that effectively demonstrate how understanding and generation enhance each other.
\end{enumerate}

This being the case, developing such a benchmark is crucial for pushing forward the research on the unification of understanding and generation, making it a promising area for future investigation.

\subsection{Multi-modal Graph Generative AI}
Graph serves as a powerful and versatile data structure used to model flexible relationships and connections between entities, being capable of modeling both naturally occurring structural \textit{instances}, e.g., protein and molecular structures, and the \textit{relations} between entities across diverse modalities, e.g., multi-modal knowledge graphs. Therefore, we introduce the concept of \textbf{Multi-modal Graph Generative AI} as a future research direction, where 1) multi-modal information can be utilized for graph generation and 2) structural relations can be used to facilitate multi-modal content generation.

\subsubsection{Leveraging multi-modal information for graph generation} Current multi-modal research predominantly focuses on modalities with regular structures with fixed degrees of freedom, e.g., texts (sequences) and images (grids). However, many real-world scenarios containing various modalities exhibit highly irregular structures with arbitrary degrees of freedom, e.g., protein structures~\cite{yi2022graph}, molecular graphs~\cite{yang2022molecule}, scene graphs~\cite{li2024scene}, etc. Accurately understanding and generating graphs across these modalities is an important direction for future research. For instance, Yao et al.~\cite{yao2024exploring} explore text-to-graph generation by leveraging the domain knowledge of LLMs, and Liu et al.~\cite{liu2024git} explore text-to-molecular graph generation by integrating the graph, image, and text information. However, there are several challenges for multi-modal graph generation: i) Understanding Structures. Given the high degree of irregularity in graphs, aligning them with various modalities poses significant difficulties. ii) Generating Structures. While mainstream approaches utilize autoregressive methods for generating discrete sequence information and employ diffusion models for generating continuous grid information, the complexity of graph structures tends to necessitate new techniques for multi-modal graph generation.

\subsubsection{Leveraging structural relations to facilitate multi-modal content generation} Traditional multi-modal learning methodologies often assume that data from different modalities are independent, whereas there can be strong intrinsic relationships across modalities in the real world~\cite{zhu2024multimodal,peng2024learning}. For example, the descriptions, chirps, and images of birds are more closely related to each other than those of other species, such as dogs and fish. Leveraging graph structure to capture these multi-modal associations may help to understand and generate new content. Ektefaie et al.~\cite{ektefaie2023multimodal} explore the combination of multiple data modalities via cross-modal dependencies and geometric relationships to develop multi-modal architectures, e.g., image-intensive, knowledge-grounded, and language-intensive models, in order to process diverse datasets. Yoon et al.~\cite{yoon2023multimodal} capture intricate relationships between multiple modalities through graphs to enhance pretrained language models with multi-modal context for generative tasks. Nevertheless, several challenges remain: i) The feature spaces of different modalities are heterogeneous, thus aligning them in a unified space via a multi-modal graph poses significant challenges. ii) The connections across instances from different modalities can be heterophilous, e.g., the meow of black and white cats may be very similar, but their visual appearances differ significantly, leading to varying degrees of weights regarding similarity for the connections across modalities within the multi-modal graph. iii) There may be substantial biases among different modalities, e.g., textual and visual modalities may dominate the learning process due to the ease of collecting texts and images via the Internet, while other modalities, such as acoustic perception and tactile sense, are much more difficult to collect.

Multi-modal graph generative AI holds significant potential applications: generating molecular graphs from texts can facilitate scientists in rapidly creating and editing chemical compounds with desired properties through natural language interactions, thereby accelerating the drug discovery process. Additionally, leveraging multi-modal graphs allows generative AI systems to reference entities associated with different modalities, thereby enhancing their ability to make cross-modal associations. Therefore, we encourage efforts in promoting future research in multi-modal graph generative AI.

\subsection{Lightweight Multi-modal Generative AI}

We define \textbf{Lightweight Multi-modal Generative AI} as the family of efficient Artificial Intelligence models capable of generating diverse types of data, including texts, images, audios, etc., while being optimized for low computational cost, fast inference, and deployment on edge devices, e.g., smartphones, IoT devices. Lightweight Multi-modal Generative AI has broad applications in various scenarios, including mobile \& edge AI, IoT \& embedded systems, and fast prototyping \& low-cost deployment.
We deem lightweight multi-modal generative AI as another promising future research direction from the following three perspectives. 


\textbf{1) Lightweight diffusion models} face challenges from sampling steps, neural architectures, and tasks. The iterative sampling process is a critical limitation of diffusion models, bringing high computational cost and constraining real-time applications. Although substantial works (e.g., distillation~\cite{sauer2023adversarial}, consistency model~\cite{song2023consistency,luo2023latent}, and flow matching~\cite{liu2023flow,lipmanflow}) engage in few-steps (e.g., 4 steps) or single-step sampling, fewer-steps sampling in general may cause remarkable quality degradation. Tasks that require high quality~\cite{tian2024emo,xu2024magicanimate} still adopt multi-step sampling. Thus, it is very important to improve the few-step sampling in future investigations. Besides, the massive network architectures of diffusion models also contribute to the issue of high computational costs, which tends to be even more severe as the model size increases rapidly. Previous methods try to obtain lightweight architectures via compression techniques such as quantization~\cite{shang2023post,tang2023post,li2023q}, pruning~\cite{zhang2024laptop}, feature cache~\cite{ma2024deepcache,chen2024delta}, and neural architecture search~\cite{tang2023lightweight,li2023autodiffusion}, etc. Although these works have achieved remarkable success, their designs are mostly tailored for the setting of multi-step sampling, either being not applicable or suffering from poor performances in few-step sampling. Therefore, exploring sampling-steps-agnostic compression methods is an important future direction as well. Moreover, traditional compression methods mainly focus on UNet-based models. Existing literature~\cite{esser2024scaling,peebles2023scalable} indicates that DiT~\cite{peebles2023scalable} may be a better architecture, resulting in the fact that more attention will be paid to DiT-based architectures. Moreover, previous compression methods mainly focus on class-condition or text-to-image generation tasks, rarely engaging in other challenging tasks such as video generation. Exploring effective compression methods for these tasks will be meaningful as well.

\textbf{2) Lightweight multi-modal LLMs}~\cite{jin2024efficient}, such as vision token compression~\cite{li2024mini,lin2023video} and efficient structures (e.g., MoE~\cite{lin2024moe} and Mamba~\cite{zhao2024cobra}), have been explored in quite a few studies. However, classic powerful compression methods (e.g., quantization and pruning) are largely unexplored for multi-modal LLM. Both diffusion models~\cite{tang2023post} and LLMs~\cite{xiao2023smoothquant} have gained successful compression rates via the utilization of quantization and pruning, giving us much confidence in exploring these methods for multi-modal LLMs in future research.

\textbf{3) Lightweight unified model} for multi-modal understanding and generation has been largely ignored in literature. However, given that the unified models typically have numerous parameters, there will be a huge need for the corresponding lightweight versions. As such, developing effective lightweight models for the unification of understanding and generation will be a frontier research direction with no doubt.

\subsection{Multi-modal Generative AI in Dynamic Environment}
The multi-modal generative models discussed so far in this paper mostly do not interact with the dynamic physical world. In the future, multi-modal generative AI agents are expected to behave like humans, where they can i) perceive the multi-modal environments, ii) conduct reasoning and planning based on the perception and their current states, iii) take action to interact with the environments, and iv) improve themselves via feedbacks from the environments. A very related topic is multi-modal embodied AI~\cite{embodied1, embodied2}, where multi-modal LLMs are used as the controller. However, existing embodied AI methods are all parameter-fixed upon deployment, limiting their abilities to self-improve in dynamic environments, where new concepts may arise in the course of time. The new concepts may cause the Out-of-Distribution (OOD) challenges for the pretrained multi-modal generative models, which fail to take the right action under these new concepts. Therefore, future works need to deal with the problem of i) when to update the model parameters, and ii) which part of the model parameters should be updated~\cite{self-directed}, e.g., the vision or the language modules.

\section{Conclusion}

In this paper, we thoroughly discuss multi-modal generative AI, with a particular focus on multi-modal LLMs, multi-modal diffusion models, as well as the unifications of LLMs and diffusions for multi-modal understanding and generation. 
We comprehensively overview two well-documented multi-modal generative AI paradigms, i.e., multi-modal LLMs for multi-modal understanding and diffusion models for visual generation. 
We deeply analyze the underlying mathematical principles, fundamental architecture designs, and practical application scenarios, indicating how these models can contribute to different aspects of multi-modal generative AI. 
We further present the necessities for the unification of understanding and generation, exploring the theoretical possibilities and potential designs towards building unified models that jointly support understanding and generation. 
The unification may come across challenges such as trade-offs between autoregressive and diffusion modeling, as well as different choices between dense and MoE architectures. 
Beyond summarizing existing methods, we also highlight promising future directions and identify the corresponding key challenges. We believe that the discussions together with the insights provided in this paper will serve as a foundation for future research and foster the development of more powerful, efficient, and generalizable multi-modal generative AI.

\scriptsize
\bibliographystyle{IEEEtran_et}
\bibliography{ref}


\normalsize

\begin{IEEEbiography}[{\includegraphics[width=1in,height=1.25in,clip,keepaspectratio]{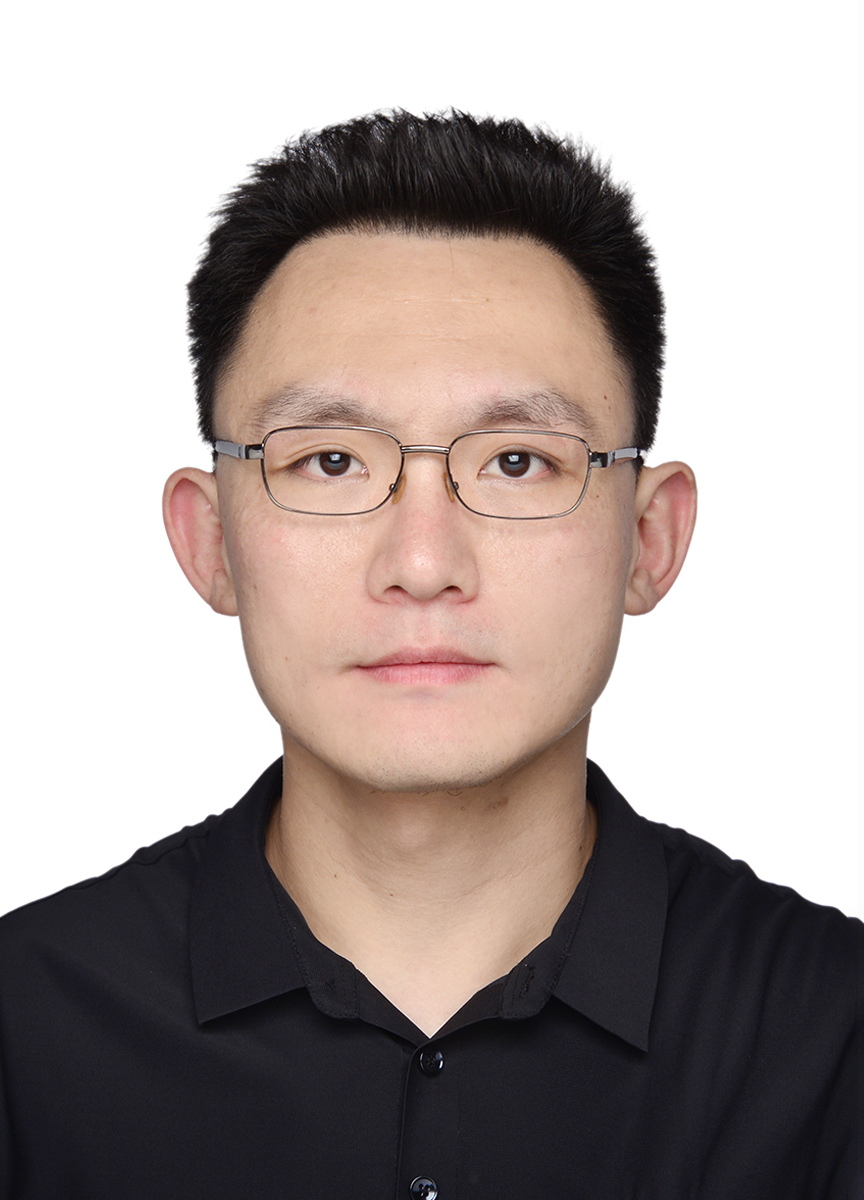}}]{Xin Wang} is currently an Associate Professor at the Department of Computer Science and Technology, Tsinghua University. He got both of his Ph.D. and B.E degrees in Computer Science and Technology from Zhejiang University, China. He also holds a Ph.D. degree in Computing Science from Simon Fraser University, Canada. His research interests include multimedia intelligence, machine learning and its applications. He has published over 200 high-quality research papers in 
top-tier conferences (ICML NeurIPS etc.) and journals (IEEE TPAMI, IEEE TIP etc.), winning three best paper awards including IEEE ICME and ACM Multimedia Asia. He is the recipient of ACM China Rising Star Award, IEEE TCMC Rising Star Award and DAMO Academy Young Fellow.
\end{IEEEbiography}

\begin{IEEEbiography}[{\includegraphics[width=1in,height=1.25in,clip,keepaspectratio]{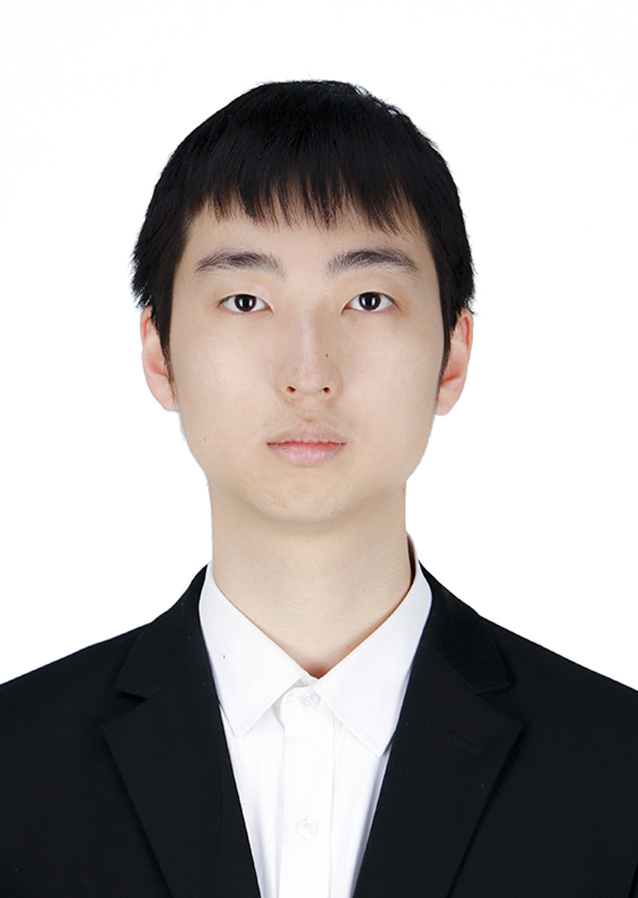}}]{Yuwei Zhou} is currently a Ph.D. student at the Department of Computer Science and Technology, Tsinghua University. He received his B.E. degree from the Department of Computer Science and Technology, Tsinghua University. His main research interests include machine learning, curriculum learning, and multi-modal generative AI.
\end{IEEEbiography}

\begin{IEEEbiography}
[{\includegraphics[width=1in,height=1.25in,clip,keepaspectratio]{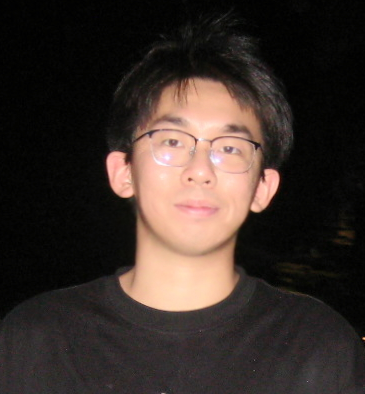}}]{Bin Huang} is currently a Ph.D. student at the Department of Computer Science and Technology, Tsinghua University. He received his B.E. degree from the Department of Computer Science and Technology, Tsinghua University. His main research interests include machine learning and multi-modal generative AI.
\end{IEEEbiography}

\begin{IEEEbiography}[{\includegraphics[width=1in,height=1.25in,clip,keepaspectratio]{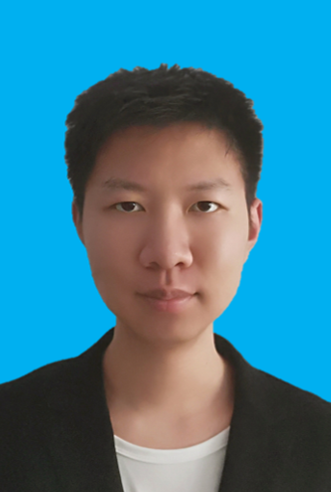}}]{Hong Chen} received B.E. from the Department of Electronic Engineering, Tsinghua University, Beijing, China in 2020. He is currently a Ph.D. candidate in the Department of Computer Science and Technology at Tsinghua University. His main research interests include machine learning, multimodal information processing.
\end{IEEEbiography}

\begin{IEEEbiography}[{\includegraphics[width=1in,height=1.25in,clip,keepaspectratio]{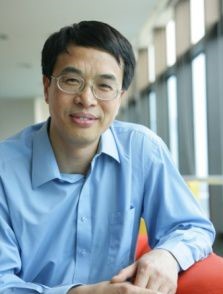}}]{Wenwu Zhu}
is currently a Professor in the Department of Computer Science and Technology at Tsinghua University. 
He received his Ph.D. degree from New York University in 1996. 
His research interests are in the area of data-driven multimedia networking and Cross-media big data computing. 
He received eight Best Paper Awards, including ACM Multimedia 2012 and IEEE TCSVT in 2001 and 2019.  
He served as EiC for IEEE TMM (2017-2019) and IEEE TCSVT (2024-2025). He served in the steering committee for IEEE TMM (2015-2016) and IEEE TMC (2007-2010), respectively. He serves as General Co-Chair for ACM Multimedia 2018 and ACM CIKM 2019, respectively. He is an AAAS Fellow, ACM Fellow, IEEE Fellow, SPIE Fellow, and a member of The Academy of Europe (Academia Europaea).
\end{IEEEbiography}


\end{document}